\pdfoutput=1

\documentclass[table]{article}

\usepackage{ACL}
\usepackage{times}
\usepackage{latexsym}

\usepackage[T1]{fontenc}

\usepackage[utf8]{inputenc}

\usepackage{microtype}

\usepackage{inconsolata}

\usepackage{graphicx}

\usepackage{xspace}

\usepackage{makecell}
\usepackage{tabularx} 
\usepackage{booktabs} 
\usepackage{marvosym}  
\usepackage{graphicx}
\usepackage{caption}
\usepackage{stfloats} 
\usepackage{cuted}
\usepackage{multirow}
\usepackage{tabularx}
\usepackage{pifont}
\usepackage{tocloft}
\usepackage{etoc}
\usepackage{pifont}
\usepackage{soul}

\usepackage{titletoc}

\newcommand{\listappendixname}{List of Appendices}
\newlistof{appendices}{apc}{\listappendixname}
\newlistof{appfigures}{afg}{List of Case Study Figures}

\usepackage[normalem]{ulem}
\useunder{\uline}{\ul}{}

\extrafloats{100}

\usepackage{listings}

\usepackage{multicol}
\usepackage{aliascnt}
\usepackage{adjustbox}

\usepackage{hypcap} 

\usepackage[most]{tcolorbox}
\usepackage{fvextra}  
\usepackage{lineno}
\usepackage{csquotes}
\usepackage[utf8]{inputenc}  
\usepackage[T1]{fontenc}     
\usepackage{graphicx}        
\usepackage{amsfonts}        
\usepackage{microtype}       
\usepackage{hyperref}        
\usepackage{url}             
\usepackage{booktabs}        
\usepackage{multirow}        
\usepackage{adjustbox}       
\usepackage{array}           
\usepackage{longtable}       
\usepackage{caption}         
\usepackage{subcaption}      
\usepackage{multirow}        
\usepackage{tcolorbox}       
\usepackage{savesym}
\usepackage{amsmath}         
\savesymbol{iint}

\usepackage{enumitem}        
\usepackage{nicematrix}      
\usepackage{pifont}          
\usepackage{makecell}        
\usepackage{mdframed}        
\usepackage[most]{tcolorbox}

\newcommand{\cmark}{\ding{51}}  
\newcommand{\xmark}{\textcolor{red}{\ding{55}}}  

\newcommand{\figcaption}{\def\@captype{figure}\caption}  
\newcommand{\tabcaption}{\def\@captype{table}\caption}    
\newcommand{\modelname}[1]{{\fontfamily{pcr}\selectfont {#1}}\xspace}
\newcommand{\benchmark}{{\fontfamily{pcr}\selectfont \textbf{R2I-Bench}}\xspace}
\newcommand{\metric}{{\fontfamily{pcr}\selectfont \textbf{R2I-Score}}\xspace}
\definecolor{mygreen}{RGB}{58, 130, 51}
\definecolor{myblue}{RGB}{40, 84, 156}
\definecolor{mygray}{RGB}{142, 142, 142}

\usepackage{amssymb}

\definecolor{commonsense_green}{HTML}{F2F8EE}   
\definecolor{mathematical_blue}{HTML}{F5F8FD}   
\definecolor{numerical_orange}{HTML}{FDF1E9}    
\definecolor{logical_purple}{HTML}{F2F2FC}      
\definecolor{concept_mixing_red}{HTML}{FEECE7}   
\definecolor{composition_yellow}{HTML}{FFF9E7}   
\definecolor{causal_rose}{HTML}{F5ECEB}         
\definecolor{backred}{RGB}{255, 190, 190}
\definecolor{backblue}{RGB}{210, 230, 250}
\definecolor{BlueGreen}{RGB}{6, 180, 185} 
\definecolor{RedOrange}{RGB}{240, 98, 51}

\newcommand{\red}[1]{$_{\color{RedOrange}\uparrow #1}$}

\newcommand{\icon}{\raisebox{-0.4em}{\includegraphics[height=1.2em]{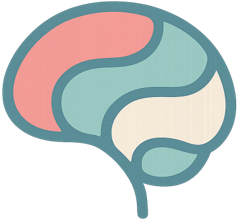}}\xspace}

\title{\icon\benchmark: Benchmarking Reasoning-Driven Text-to-Image Generation}

\author{Kaijie Chen{\footnotesize $^{\bigstar}$}, Zihao Lin$^{*}${\footnotesize $^{\spadesuit}$}, Zhiyang Xu$^{*}${\footnotesize $^{\clubsuit}$}, Ying Shen$^{*}${\footnotesize $^{\blacklozenge}$}, \\
\textbf{Yuguang Yao{\footnotesize $^{\heartsuit}$}, Joy Rimchala{\footnotesize $^{\diamondsuit}$}, Jiaxin Zhang{\footnotesize $^{\diamondsuit}$}, Lifu Huang{\footnotesize $^{\spadesuit}$$^{\textsuperscript{\Letter}}$}}\\
}

\begin{document}

\twocolumn[{%
    \renewcommand\twocolumn[1][]{#1}
    \maketitle
    \centering
   \vspace{-42pt}
    \vspace{1em}
    \begin{center}
        \centering
        \includegraphics[width=\textwidth]{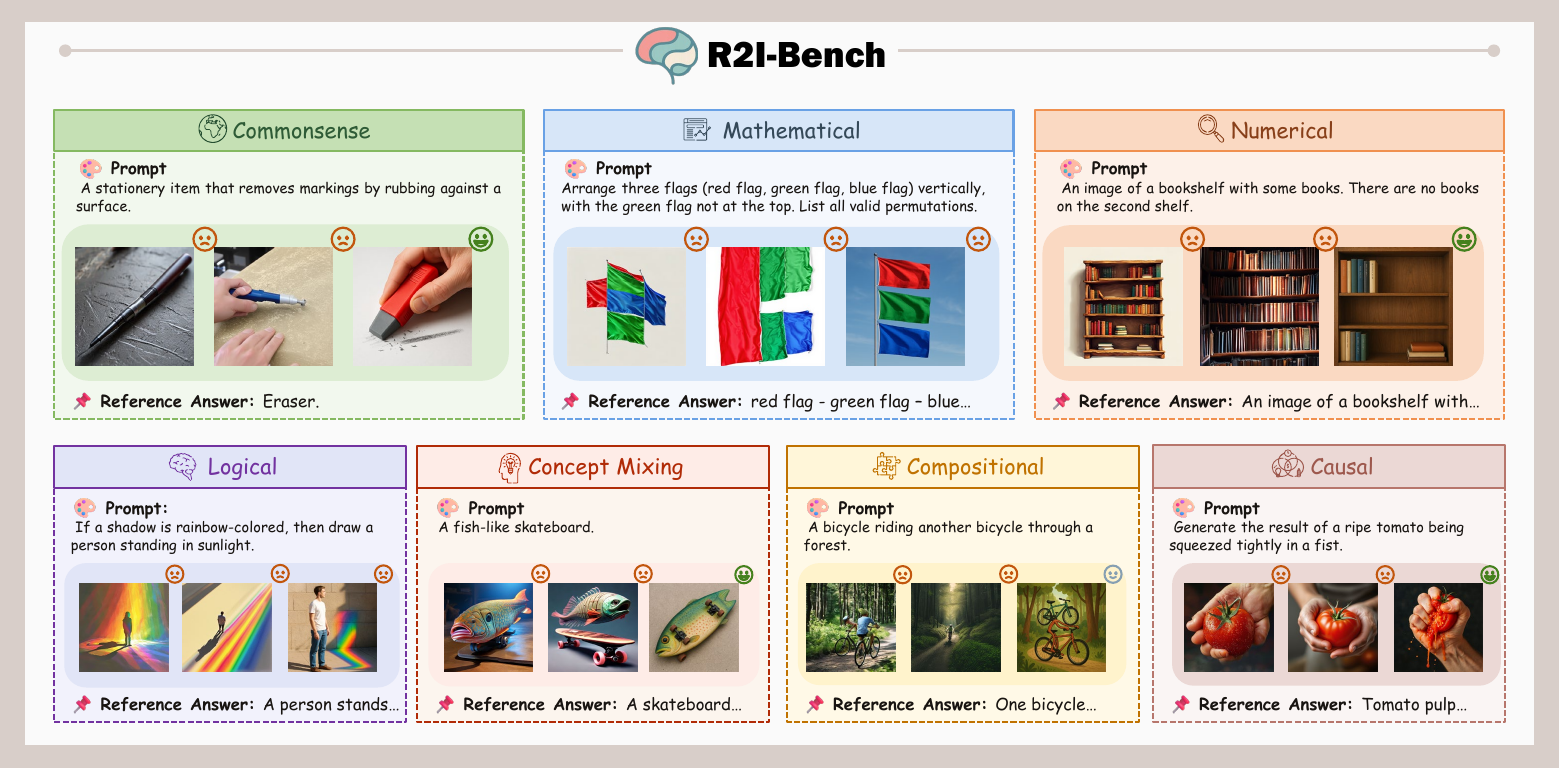}
        \vspace{-7mm}
        \captionof{figure}{We introduce \textbf{\benchmark}, a comprehensive benchmark designed to assess the reasoning capabilities of text-to-image (T2I) generation models. It encompasses $7$ primary reasoning categories, which are further subdivided into $32$ fine-grained subcategories.}
        \vspace{2ex}
        \label{fig:main}

    \end{center}
}
]

\author{Kaijie Chen{\footnotesize $^{\bigstar}$}, Zihao Lin$^{*}${\footnotesize $^{\spadesuit}$}, Zhiyang Xu$^{*}${\footnotesize $^{\clubsuit}$}, Ying Shen$^{*}${\footnotesize $^{\blacklozenge}$}, \\
\textbf{Yuguang Yao{\footnotesize $^{\heartsuit}$}, Joy Rimchala{\footnotesize $^{\diamondsuit}$}, Jiaxin Zhang{\footnotesize $^{\diamondsuit}$}, Lifu Huang{\footnotesize $^{\spadesuit}$$^{\textsuperscript{\Letter}}$}}\\
}
{\let\thefootnote\relax\footnote{$^{*}$~Equal Contribution $^{\bigstar}$~Tongji University $^{\spadesuit}$~University of California, Davis $^{\clubsuit}$~Virginia Tech {\footnotesize $^{\blacklozenge}$}~UIUC {\footnotesize $^{\heartsuit}$}~Michigan State University $^{\diamondsuit}$~Intuit AI Research }}

\begin{abstract}

Reasoning is a fundamental capability often required in real-world text-to-image (T2I) generation, e.g., generating ``\textit{a bitten apple that has been left in the air for more than a week}'' necessitates understanding temporal decay and commonsense concepts. While recent T2I models have made impressive progress in producing photorealistic images, their reasoning capability remains underdeveloped and insufficiently evaluated. To bridge this gap, we introduce \benchmark, a comprehensive benchmark specifically designed to rigorously assess reasoning-driven T2I generation. 
\benchmark comprises $3,068$ meticulously curated data instances, spanning $7$ core reasoning categories, including commonsense, mathematical, logical, compositional, numerical, causal, and concept mixing. To facilitate fine-grained evaluation, we design \metric, a QA-style metric based on instance-specific, reasoning-oriented evaluation questions that assess three critical dimensions: \textit{text-image alignment}, \textit{reasoning accuracy}, and \textit{image quality}. Extensive experiments with 16 representative T2I models, including a strong pipeline-based framework that decouples reasoning and generation using the state-of-the-art language and image generation models, demonstrate consistently limited reasoning performance, highlighting the need for more robust, reasoning-aware architectures in the next generation of T2I systems. Project page: \url{https://r2i-bench.github.io}.

\end{abstract}

\section{Introduction}
\label{sec:intro}
\begin{table*}[htbp]
\vspace{-8mm}
\centering

\resizebox{0.95\textwidth}{!}{  
\begin{tabular}{@{}lccccccccc@{}}  
\toprule
\multirow{2}{*}{\bf Benchmarks} & \multicolumn{7}{c}{\bf Reasoning Capabilities Evaluated in Text-to-Image Generation} & \multirow{2}{*}{\bf Human} \\ \cmidrule{2-8}
\cmidrule(l){2-3} \cmidrule(l){4-8} 
 & {\footnotesize \cellcolor{commonsense_green}\textbf{Commonsense}} & {\footnotesize \cellcolor{composition_yellow}\textbf{Compositional}} & {\footnotesize \cellcolor{numerical_orange}\textbf{Numerical}} & {\footnotesize \cellcolor{mathematical_blue}\textbf{Mathematical}} & {\footnotesize \cellcolor{concept_mixing_red}\textbf{Concept Mixing}} & {\footnotesize \cellcolor{logical_purple}\textbf{Logical}} & {\footnotesize \cellcolor{causal_rose}\textbf{Causal}} & {\bf Annotation}\\ 
\midrule
OK-VQA~\cite{marino2019ok} & \cmark & \xmark & \xmark & \xmark & \xmark & \xmark & \xmark & \xmark \\ 
Winoground~\cite{thrush2022winoground} & \xmark & \cmark & \xmark & \xmark & \xmark & \xmark & \xmark & \xmark \\ 
HEIM~\cite{lee2023holistic} & \xmark & \cmark & \xmark & \xmark & \xmark & \xmark & \xmark & \xmark \\ 
GeckoNum~\cite{ghosh2023geneval} & \xmark & \xmark & \cmark & \xmark & \xmark & \xmark & \xmark & \cmark \\ 
GenEval~\cite{ghosh2023geneval} & \xmark & \cmark & \cmark & \xmark & \xmark & \xmark & \xmark & \xmark \\ 
GenAI-Bench~\cite{li2024genai} & \xmark & \xmark & \cmark & \xmark & \xmark & \xmark & \xmark & \xmark \\ 
ConceptMix~\cite{wu2024conceptmix} & \xmark & \cmark & \xmark & \xmark & \cmark & \xmark & \xmark & \xmark \\ 
Commonsense-T2I~\cite{fu2024commonsense} & \cmark & \xmark & \xmark & \xmark & \xmark & \xmark & \xmark & \xmark \\ 
WISE~\cite{niu2025wise} & \cmark & \xmark & \xmark & \xmark & \xmark & \xmark & \xmark & \xmark \\ 

\midrule 
{\bf \benchmark (Ours)} & \cellcolor{commonsense_green}\cmark & \cellcolor{composition_yellow}\cmark & \cellcolor{numerical_orange}\cmark & \cellcolor{mathematical_blue}\cmark & \cellcolor{concept_mixing_red}\cmark & \cellcolor{logical_purple}\cmark & \cellcolor{causal_rose}\cmark  & \cmark \\ 
\bottomrule
\end{tabular}
}
\vspace{-2mm}
\caption{{\bf Comparison between \benchmark and existing text-to-image benchmarks.} \benchmark covers a broader spectrum of essential reasoning capabilities for text-to-image generation. In addition, \benchmark provides manually curated, high-quality evaluation criteria to support rigorous and consistent assessment. }
\label{tab:skill_comparison}
\vspace{-5mm}  
\end{table*}

Reasoning is a fundamental capability underpinning most, if not all, human cognitive tasks, including text-to-image (T2I) generation. In real-world scenarios, prompts often require models to go beyond surface-level descriptions and engage in multi-step reasoning. For example, generating an image for ``\textit{a bitten apple that has been left in the air for more than a week}'' requires understanding the concept of decay over time, inferring the visual appearance of a spoiled apple, composing that with contextual cues, and finally generating an image to depict ``\textit{a bitten and spoiled apple}''. 

However, despite recent advances, most existing T2I models, whether based on diffusion~\cite{esser2024scaling, xie2025sana15efficientscaling, qin2025lumina, yang2024improvingdiffusionbasedimagesynthesis}, autoregressive transformer~\cite{sun2024autoregressive, zhang2024var, chen2025janus, wang2024emu3, chen2024spark}, or unified architectures\cite{xiao2024omnigen, xie2024show, zhou2024transfusion, tong2024metamorph, sun2023emu, sun2024generativemultimodalmodelsincontext}, primarily focus on \textit{semantic rendering}, where the prompt explicitly specifies what to generate and the model simply converts it into an image. Although recent work~\cite{jiang2025t2i, guo2025can, liao2025imagegen} has begun to benchmark and enhance reasoning-driven T2I generation, they are often limited to narrow domains such as commonsense \cite{niu2025wise}, numerical reasoning \cite{ghosh2023geneval}, or concept mixing \cite{wu2024conceptmix}. Furthermore, widely adopted evaluation metrics for T2I generation, such as CLIPScore~\cite{hessel2021clipscore}, VQAScore~\cite{lin2024evaluating}, and WIScore~\cite{niu2025wise}, mainly assess the semantic alignment between generated images and prompts or fail to generalize across diverse reasoning types, limiting meaningful development, comparison, and assessment of the underlying reasoning capabilities in T2I generation models.

To bridge these gaps, we introduce \textbf{\benchmark} (\textbf{R}easoning-\textbf{to}-\textbf{I}mage \textbf{Bench}mark), a comprehensive benchmark consisting of $3,068$ meticulously curated text prompts, specifically designed to evaluate the reasoning capabilities of T2I models. Each prompt is initially generated using a state-of-the-art large language model (i.e., \modelname{GPT-4o}) and subsequently validated and refined by domain experts to ensure the quality and reliability. As shown in Figure~\ref{fig:main},  
\benchmark encompasses $7$ core reasoning categories, including \textbf{\textit{commonsense}}, \textbf{\textit{compositional}}, \textbf{\textit{logical}}, \textbf{\textit{mathematical}}, \textbf{\textit{causal}}, \textbf{\textit{numerical}}, and \textbf{\textit{concept mixing}}, which are further subdivided into $32$ fine-grained reasoning subcategories. In contrast to prior T2I evaluation datasets, \benchmark offers significantly broader and more systematic coverage of diverse reasoning skills, as summarized in Table~\ref{tab:skill_comparison}.

To enable fine-grained evaluation of reasoning-driven T2I generation, each T2I prompt in \benchmark is paired with a set of instance-specific diagnostic questions and corresponding scoring criteria, all verified by human experts. These questions assess the quality of T2I generation along three critical aspects: (1) text-image alignment, (2) reasoning accuracy, and (3) image quality. Building on these evaluation questions and criteria, we introduce a QA-style metric, \textbf{\metric}, which aggregates scores using a weighted scheme. \textbf{\metric} demonstrates strong alignment with human judgments, offering a more faithful and interpretable performance measure of T2I models on \benchmark.

\begin{figure*}[t!]

\centering
\includegraphics[width=0.95\textwidth]{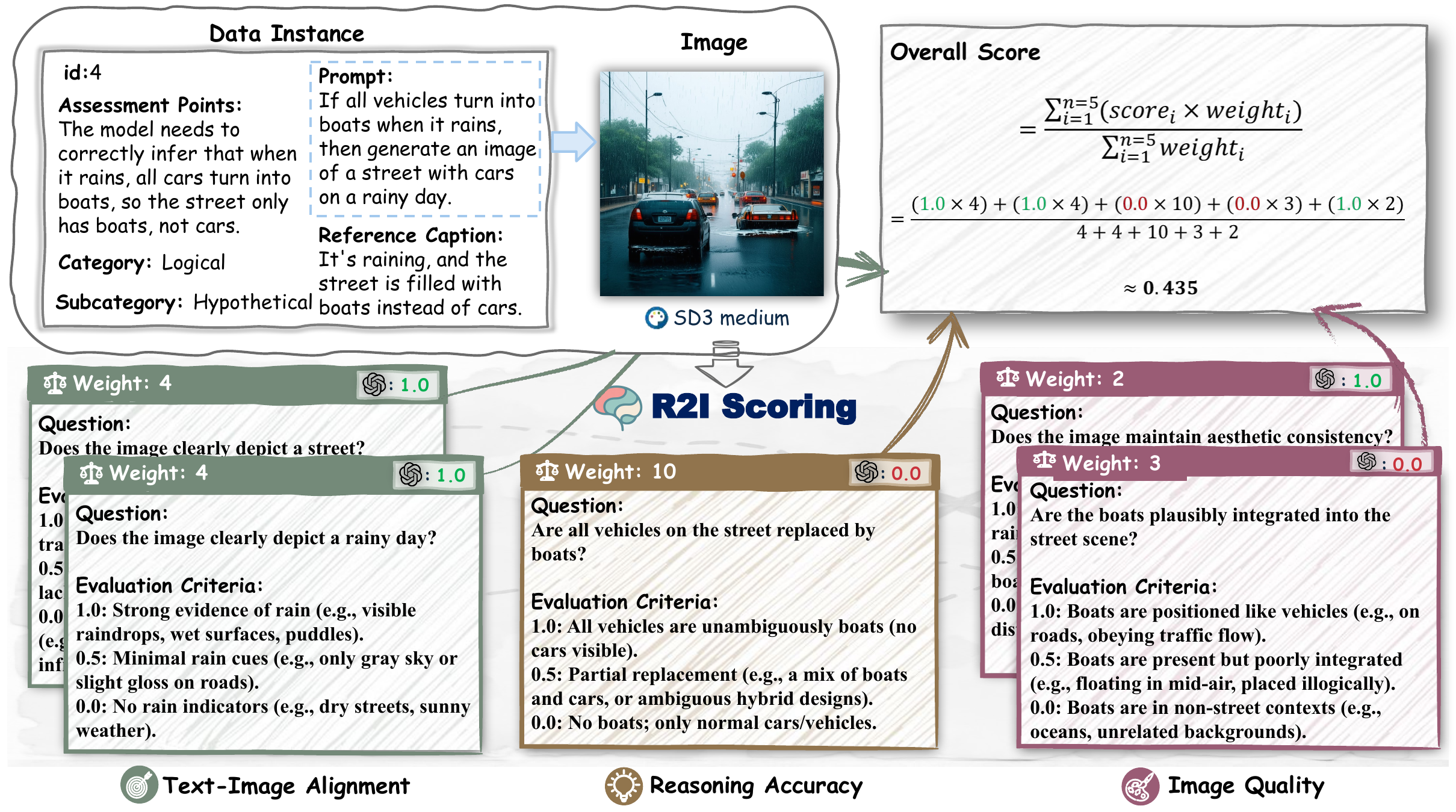}

\caption{Example Illustration of \benchmark and \metric.}
\label{fig:mainevaluation}

\end{figure*}

We systematically evaluate 16 representative T2I models on \benchmark, spanning diffusion-based, autoregressive, reasoning-enhanced, and closed-source models. To further explore the upper bound of reasoning-driven T2I generation, we also develop a strong pipeline-based framework that decouples reasoning and generation: a state-of-the-art LLM (\modelname{GPT-4o}) first performs reasoning over the prompt and rewrites it into a detailed description, which is then rendered by a high-performing T2I model (\modelname{SD3-medium}). Experimental results reveal several key insights: \textbf{(1)} All the open-source models achieve less than $45\%$ accuracy, demonstrating limited reasoning capabilities in existing T2I models and underscoring the significance of \benchmark as a rigorous evaluation benchmark. Notably, these models tend to interpret prompts as bags of words, e.g., they generate both objects for the prompt ``\textit{either a spoon or a bowl}'', disregarding the logical disjunction; \textbf{(2)} Mathematical reasoning remains a persistent challenge across all models, largely due to the lack of diverse, high-quality training data grounded in mathematical concepts and their visual representations; \textbf{(3)} Recent efforts to enhance reasoning through Chain of Thought (CoT) or Reinforcement Learning (RL)~\cite{guo2025can, liao2025imagegen, jiang2025t2i} yield marginal improvements, highlighting the need for more robust, fundamentally reasoning-aware T2I models; and \textbf{(4)} While the pipeline-based framework improves performance, it still struggles with abstract mathematical reasoning and accurately interpreting specific linguistic constructs such as quantities, limiters, and quantifiers. Finally, we also conduct a comprehensive qualitative error analysis, categorizing model failures into three main categories, including reasoning errors, visual element errors, and image quality degradation, providing valuable insights to future research.

Our contributions are summarized as follows: (1) We introduce \benchmark, the first comprehensive benchmark specifically designed to evaluate reasoning-driven T2I generation. Covering a broad range of reasoning categories and meticulously curated through a rigorous human-in-the-loop process, \benchmark offers a valuable resource for benchmarking and advancing T2I models. (2) To enable fine-grained evaluation of reasoning-driven T2I generation, we design a new metric, \textbf{\metric}, built on human-validated evaluation questions and scoring criteria tailored to each data instance in \benchmark. \textbf{\metric} assesses model performance across three critical dimensions, including text-image alignment, reasoning accuracy, and image quality. (3) Through extensive experiments and analysis, we identify several key limitations in all the existing T2I models and provide valuable insights for future research.

\section{Related Work}
\label{sec:relatedworks}

\vspace{-2mm}
\paragraph{Text-to-Image Generation Models}

Recent advances in text-to-image (T2I) generation have produced high-quality models across various architectures, including diffusion~\cite{esser2024scaling, gao2024lumina, xie2025sana15efficientscaling, qin2025lumina, yang2024improvingdiffusionbasedimagesynthesis}, autoregressive~\cite{sun2024autoregressive, zhang2024var, chen2025janus, wang2024emu3}, and unified frameworks~\cite{xiao2024omnigen, xie2024show, zhou2024transfusion, tong2024metamorph, sun2023emu, sun2024generativemultimodalmodelsincontext}. 
More recently, reasoning-augmented models have incorporated chain-of-thought (CoT) reasoning~\cite{liao2025imagegen} and reinforcement learning~\cite{guo2025can,jiang2025t2i} to better handle complex prompts. However, their reasoning capability remains underdeveloped and insufficiently evaluated.

\vspace{-2mm}
\paragraph{Text-to-Image Evaluation Benchmarks and Metrics}
Existing T2I benchmarks evaluate isolated reasoning skills but lack comprehensive coverage. OK-VQA~\cite{marino2019ok}, WISE~\cite{niu2025wise}, and Commonsense T2I~\cite{fu2024commonsense} emphasize shallow or knowledge-based reasoning, while GeckoNum~\cite{kajic2024evaluating} focuses solely on numerical tasks. Benchmarks like Winoground~\cite{thrush2022winoground}, GenEval~\cite{ghosh2023geneval}, and GenAI-Bench~\cite{li2024genai} target compositionally.  

Despite progress, no existing benchmark offers a unified framework for evaluating the full spectrum of T2I reasoning abilities (see Table~\ref{tab:skill_comparison}). 
Current evaluation metrics also lack reasoning sensitivity. CLIPScore~\cite{hessel2021clipscore}, DSGScore~\cite{cho2023davidsonian}, and VQAScore~\cite{lin2024evaluating} underperform on complex reasoning and struggle with compositional or numerical fidelity. LLM-based metrics such as LLMScore~\cite{lu2023llmscore} and SemVarEffect~\cite{zhu2024evaluating} overlook spatial or relational accuracy. While RIScore~\cite{zhao2025envisioning} and WIScore~\cite{niu2025wise} offer GPT-based scoring, they lack the granularity needed for fine-grained evaluation. Thus, a critical gap remains in metrics that rigorously assess reasoning in T2I generation.

\section{\benchmark}
\vspace{-2mm}
\paragraph{Overview} As shown in the top left part of Figure~\ref{fig:mainevaluation}, each data instance in \benchmark consists of four elements: (1) a reasoning-based T2I prompt which serves as a textual input to the T2I models; (2) a reference caption that explicitly describes the content of the image that is supposed to be generated; (3) an explanation description, which explains the reasoning steps from the T2I prompt to the reference caption and is used to generate reasoning-driven evaluation questions; and (4) the category, the subcategory,  and the index of the data instance. As illustrated in Appendix~\ref{sec:data-generation-pipeline} Figure~\ref{fig:dataconstruction}, we adopt a human-in-the-loop pipeline to construct \benchmark, which comprises three main stages: (1) data collection, (2) data filtering, and (3) evaluation criteria generation.

\vspace{-2mm}
\paragraph{Data Collection}
In the initial stage, a team of five human experts systematically reviews prior work relevant to text-to-image (T2I) reasoning tasks~\cite{wu2024conceptmix, kajic2024evaluating, thrush2022winoground, li2024genai, liu2020logiqa, liew2022magicmix, fu2024commonsense, chevalley2022causalbench, niu2025wise, lei2025imagine, lee2023holistic}. Based on this comprehensive analysis, they identify $7$ core reasoning categories frequently required across diverse T2I scenarios: \textbf{\textit{commonsense}}, \textbf{\textit{compositional}}, \textbf{\textit{logical}}, \textbf{\textit{concept-mixing}}, \textbf{\textit{numerical}}, \textbf{\textit{mathematical}}, and \textbf{\textit{causal reasoning}}. These primary categories are further refined into $32$ fine-grained subcategories, as illustrated in Figure~\ref{fig:sunburnt}. Detailed definitions for all the core and fine-grained reasoning categories are provided in Appendix~\ref{sec:category_definitions}.

For each subcategory, we instruct \modelname{GPT-4o} to generate $100\text{-}120$ T2I prompts designed to test the corresponding reasoning skill, accompanied by reference captions for subsequent evaluation. To ensure that the prompts emphasize reasoning and avoid direct visual descriptions, the generation instruction is constrained by two key guidelines: (1) prompts must not explicitly reveal the answer or directly describe visual features, and (2) the corresponding visual elements must be uniquely identifiable. In-context learning is used, where the model is conditioned on three positive and three negative exemplar prompts authored by human experts. For each generated instance, based on the T2I prompt and reference caption, we further instruct \modelname{GPT-4o} to generate an explanation description which will be leveraged to generate evaluation criteria in the later stage. The instructions for generating the T2I prompt and explanation description are shown in Appendix~\ref{sec:prompt_creation} and \ref{sec:explanation_creation} respectively. 

\vspace{-2mm}
\paragraph{Data Filtering and Refinement}
To ensure the quality and validity of the collected data instances, we conduct manual filtering to exclude instances where the prompt fails to yield a renderable image or the associated visual elements are not uniquely identifiable. This filtering step yields approximately 800 high-quality instances from the initial set of 3,200.
To expand the dataset while preserving both diversity and quality, we treat these 800 instances as seed T2I prompts. Human experts then engage in an iterative refinement process with \modelname{GPT-4o}, prompting the model to generate additional candidates. After each generation round, human experts evaluate the generated T2I prompts and provide targeted feedback to guide revisions, ensuring that each prompt adheres to the two guidelines. This iterative augmentation continues until each reasoning subcategory reaches approximately 100 validated instances.

\begin{figure}
    \centering
    \includegraphics[width=\linewidth]{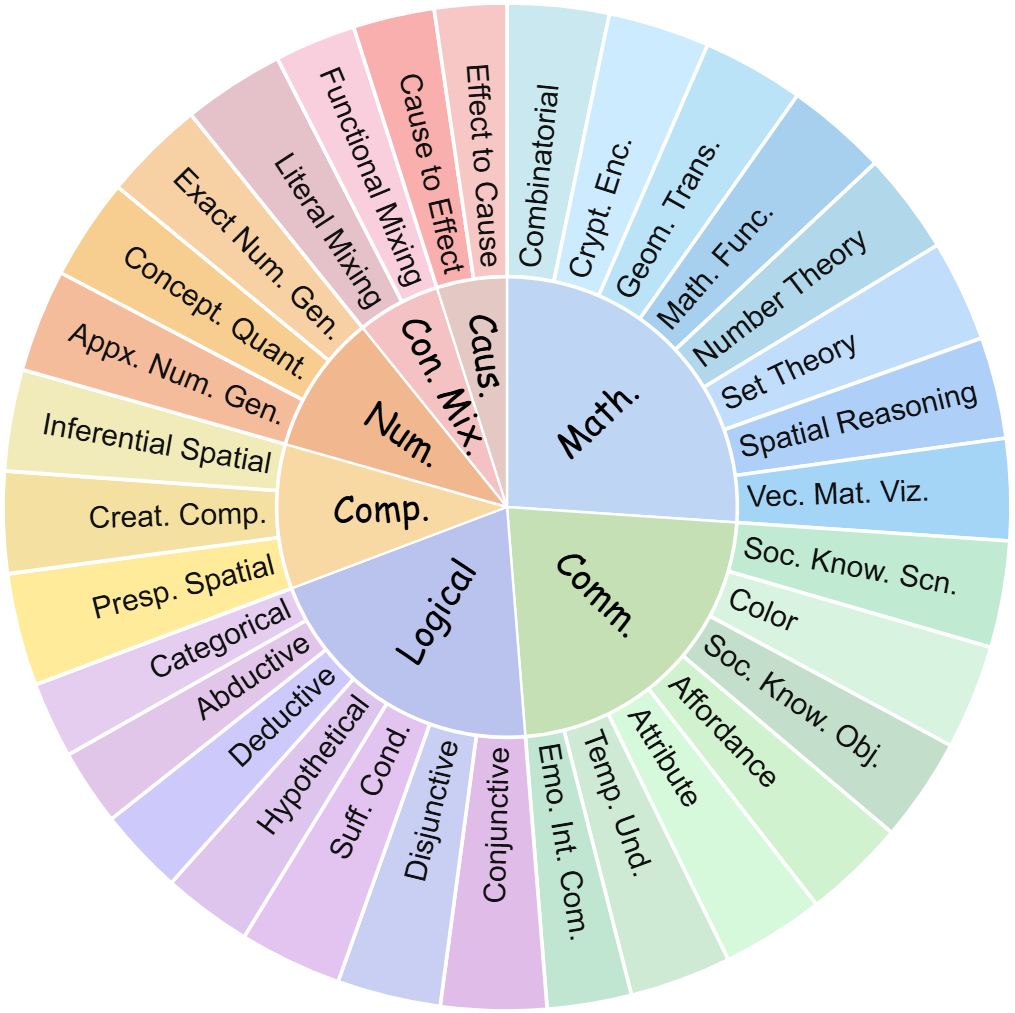}
\caption{\textbf{Distribution of Diverse Reasoning Categories in \benchmark.} \underline{Caus.}: Causal. \underline{Con. Mix.}: Concept Mixing. \underline{Math.}: Mathematical. \underline{Comm.}: Commonsense. \underline{Num.}: Numerical. \underline{Comp.}: Compositional.
 }
\label{fig:sunburnt}
\vspace{-6mm}
\end{figure}

\vspace{-2mm}
\paragraph{Evaluation Criteria Generation and \metric}
Existing T2I evaluation metrics often fail to adequately assess the reasoning abilities essential for high-quality image generation. Hence, we create an evaluation set (i.e., a set of evaluation questions and their corresponding scoring criteria) tailored to each data instance in \benchmark.  The carefully designed evaluation questions assess the T2I models in three core dimensions: \textit{\ding{172} Text-image alignment}: whether the generated image accurately contains all required elements, such as objects and attributes; \textit{\ding{173} Reasoning accuracy}: whether the T2I model performs necessary reasoning over the input prompt to correctly generate the output image; \textit{\ding{174} Image quality}: measuring the clarity and quality (e.g., vagueness, distortions, and so on) of the generated images. Example questions for each evaluation dimension are provided in Figure~\ref{fig:mainevaluation}.

For efficient, we feed each previously generated T2I prompt,  the corresponding reference caption, and explanation description to \modelname{GPT-4o} and ask it to generate a set of evaluation questions, each paired with an assigned evaluation dimension, an importance weight, and a scoring criterion. To further emphasize reasoning over surface-level features, we manually set a weight constraint range for each question based on its evaluation dimension: $[7,10]$ for \textit{reasoning accuracy}, $[4,6]$ for \textit{text-image alignment}, and $[1,3]$ for \textit{image quality}. This design reflects our goal of benchmarking reasoning-driven T2I generation, under the assumption that most modern T2I models already perform well in producing visually appealing images. To ensure reliability and consistency, all evaluation questions, scoring criteria, and importance weights are manually validated and refined by expert annotators. The complete instruction template in this process are provided in Appendix~\ref{sec:promptevaluationgen}. 

Building on the evaluation set, we propose a new QA-style metric, \metric. Given a generated image for a T2I prompt, we feed the image along with each evaluation question and its corresponding scoring criteria as input to \modelname{GPT-4o}, and ask it to select a score $s_i$ based on the provided criteria. This scoring instruction template is detailed in Appendix~\ref{sec:promptimgevaluation}. We calculate \metric as follows:
\begin{equation}
    \text{R2I-Score} = \frac{\sum_{i=1}^{n} w_i \cdot s_i}{\sum_{i=1}^{n} w_i}
\end{equation}
where $n$ is the total number of evaluation questions for a given instance, and $w_i$ is the importance weight assigned to the $i$th evaluation question.

\vspace{-2mm}
\paragraph{Dataset Statistics} Finally, \benchmark comprises $3,068$ high-quality, reason-driven T2I prompts. Figure~\ref{fig:main} includes an example T2I prompt for each core reasoning category. Figure~\ref{fig:sunburnt} presents the distribution of these categories in \benchmark, and Table~\ref{tab:key_statistics} in Appendix \ref{sec:stat-of-benchmark} provides detailed statistics for \benchmark.

\begin{table*}[t]
 
\centering
\small  
\resizebox{0.99\textwidth}{!}{
\begin{tabularx}{\textwidth}{
  >{\raggedright\arraybackslash}p{2.15cm}  
  >{\centering\arraybackslash}p{0.6cm} 
  >{\centering\arraybackslash}p{0.7cm}
  >{\centering\arraybackslash}p{1.6cm}
  >{\centering\arraybackslash}p{1.6cm}
  >{\centering\arraybackslash}p{0.9cm}
  >{\centering\arraybackslash}p{0.75cm}
  >{\centering\arraybackslash}p{1.1cm}
  >{\centering\arraybackslash}p{1.5cm}
  >{\centering\arraybackslash}p{0.9cm}
}
\toprule

Model & 
Size & 
Overall & 
\cellcolor{commonsense_green}Commonsense & 
\cellcolor{composition_yellow}Compositional& 
\cellcolor{concept_mixing_red}Con.Mix.& 
\cellcolor{logical_purple}Logical & 
\cellcolor{numerical_orange}Numerical & 
\cellcolor{mathematical_blue}Mathematical& 
\cellcolor{causal_rose}Causal \\
\midrule
\multicolumn{10}{c}{\textit{Diffusion Models}} \\
\cmidrule{1-10} 
SD3-medium  & 2B &\colorbox{backblue!75}{0.45} & 0.54 & 0.64 & 0.63 & 0.55 & 0.50 & \colorbox{backblue!75}{0.19} & 0.18 \\
\fontsize{8.3}{0}\selectfont{Lumina-Image 2.0} & 2.6B & 0.42 & 0.49 & 0.65 & 0.54 & \colorbox{backblue!75}{0.56} & 0.43 & 0.13 & 0.40 \\
Sana-1.5 & 4.8B & 0.41 & 0.49 & \colorbox{backblue!75}{0.67} & \colorbox{backblue!75}{0.66} & 0.49 & 0.48 & 0.13 & 0.21 \\
Lumina-T2I & 5B & 0.33 & 0.38 & 0.49 & 0.55 & 0.38 & 0.45 & 0.13 & 0.18 \\
Omnigen & 3.8B & 0.40 & 0.43 & 0.60 & 0.43 & 0.51 & 0.47 & 0.18 & 0.34 \\
LLM4GEN\textit{$_{\text{SD1.5}}$}  & 0.86B & 0.40 & \colorbox{backblue!75}{0.55} & 0.48 & 0.60 & 0.55 & 0.39 & 0.07 & \colorbox{backblue!75}{0.45} \\
ELLA\textit{$_{\text{SD1.5}}$} & 0.07B & 0.31 & 0.40 & 0.44 & 0.40 & 0.40 & 0.32 & 0.07 & 0.29 \\
\cmidrule{1-10}
\multicolumn{10}{c}{\textit{AutoRegressive Models}} \\
\cmidrule{1-10}
EMU3 & 8.0B & 0.41 & 0.44 & 0.59 & 0.62 & 0.55 & \colorbox{backblue!75}{0.61} & 0.09 & 0.41 \\
Janus-Pro-7B & 7B & 0.38 & 0.45 & 0.60 & 0.64 & 0.46 & 0.46 & 0.07 & 0.36 \\
LlamaGen & 0.8B & 0.29 & 0.38 & 0.39 & 0.49 & 0.38 & 0.35 & 0.07 & 0.12 \\
Show-o & 1.3B & 0.36 & 0.42 & 0.59 & 0.56 & 0.42 & 0.57 & 0.12 & 0.30 \\
\cmidrule{1-10}
\multicolumn{10}{c}{\textit{Reasoning-Enhanced Models}} \\
\cmidrule{1-10}
Show-o+ORM & 1.3B & 0.34  &  0.42 & 0.45  & 0.44  &  0.37 & 0.49  & 0.12  & 0.26  \\
Show-o+DPO & 1.3B & 0.36 &  0.43 &  0.47 &  0.48 &  0.41 & 0.51 & 0.13  &  0.31 \\
Show-o+PARM & 1.3B & 0.38 & 0.45 & 0.49 & 0.51 & 0.45 & 0.56 & 0.13 & 0.32 \\
\cmidrule{1-10} 
\multicolumn{10}{c}{\textit{Close Source Models}} \\
\cmidrule{1-10}
DALL-E-3 & - & 0.71 & 0.78 & 0.76 & 0.86 & 0.69 & 0.69 & 0.21 & 0.64 \\
gpt-image-1 & - & \colorbox{backred!50}{0.77} & \colorbox{backred!50}{0.83} & \colorbox{backred!50}{0.87} & \colorbox{backred!50}{0.89} & \colorbox{backred!50}{0.81} & \colorbox{backred!50}{0.88} & \colorbox{backred!50}{0.58} & \colorbox{backred!50}{0.71} \\
\cmidrule{1-10}
\multicolumn{10}{c}{\textit{Prompt-Rewrite Pipeline}} \\
\cmidrule{1-10}
\fontsize{7.1}{10}\selectfont{gpt-4o+SD3-medium} & 2B & 0.58\red{0.13}  &  0.75\red{0.21} & 0.75\red{0.11}  & 0.81\red{0.18}  &  0.65\red{0.10} & 0.63\red{0.13} & 0.22\red{0.03} & 0.76\red{0.58}  \\
\cmidrule{1-10}
\end{tabularx}
}
\vspace{-3mm}
\caption{\textbf{Evaluation on \benchmark.} The highest accuracy for \colorbox{backred!50}{closed-source} and \colorbox{backblue!75}{open-source} text-to-image models are marked in red and blue respectively. Con.Mix.: Concept Mixing.
}
\label{tab:main_table}

\end{table*}

\section{Experiments}  

\subsection{Experimental Setup} \label{sec:setup}

To conduct evaluation on \benchmark, we carefully select 16 representative, high-performing T2I models with publicly available model checkpoints, spanning four distinct categories: 

(1) \textit{Diffusion Models}, featuring models including SD3-medium~\cite{rombach2022high},
Lumina-Image 2.0~\cite{qin2025lumina},
Sana-1.5~\cite{xie2025sana15efficientscaling},
Lumina-T2I~\cite{lumina2},
Omnigen~\cite{xiao2024omnigen},
LLM4GEN\textit{$_{\text{SD1.5}}$}~\cite{liu2025llm4gen},
and ELLA\textit{$_{\text{SD1.5}}$}~\cite{hu2024ella};
(2) \textit{Autoregressive Models}, including EMU3~\cite{wang2024emu3},
Janus-Pro-7B~\cite{chen2025janus},
LlamaGen~\cite{sun2024autoregressive},
and Show-o~\cite{xie2024show}; 
(3) \textit{Reasoning-Enhanced Models}, including Show-o+ORM, Show-o+DPO, and Show-o+PARM~\cite{guo2025can}
; 
and (4) \textit{Closed-Source Models}, including DALL-E-3~\cite{ma2024learning} 
and gpt-image-1~\cite{hurst2024gpt}.
Additional implementation details, such as model architectures, configurations, and inference parameters, are provided in Appendix~\ref{appendix:model-config}. For evaluation, we adopt the proposed \metric metric.

Intuitively, reasoning-driven T2I generation could be more effectively addressed by decoupling reasoning from image generation—first leveraging a large language model to perform complex reasoning and generate a detailed textual description, and then using a powerful image generation model to render the final image~\cite{niu2025wise}. Motivated by this, we design a strong pipeline-based framework that explicitly separates the reasoning and generation stages.

The framework first employs a state-of-the-art LLM (\modelname{GPT-4o}) to interpret and reason over the original prompt, producing a detailed and structured image description. This rewritten prompt is then passed to a high-performing T2I model (\modelname{SD3-medium}) to generate the corresponding image. We name this pipelined framework as \modelname{gpt-4o+SD3-medium}.

\subsection{Main Results} \label{sec:exp_analysis}

Table~\ref{tab:main_table} presents the evaluation results of all T2I models across the core reasoning categories in \benchmark, with detailed subcategory-level results provided in Appendix~\ref{sec:detailresults}. The main findings are summarized as follows.

\vspace{-2mm}
\paragraph{T2I Models Show Limited Capability in Reasoning-Driven Image Generation.} Our evaluation reveals that most open-source models achieve a score lower than 45\% based on \metric, suggesting a notable gap in their ability to handle reasoning-driven T2I prompts. This limitation appears to stem from a shallow understanding of prompts, often interpreted as a bag of words rather than through compositional or logical reasoning. This hypothesis is further supported by our qualitative error analysis, illustrated in Appendix~\ref{sec:visualization_subcategory}, Figures~\ref{fig:appendix_commonsense} through~\ref{fig:appendix_causal}, where the majority of models simply generate images that merely reflect the objects explicitly mentioned in the prompt without performing necessary inferential reasoning. 

For instance, given the prompt ``\textit{a cat-like bed}'' (Figure~\ref{fig:appendix_concept_mixing}), most of the models, including \modelname{EMU3}, \modelname{SD3-medium}, \modelname{ELLA}, and \modelname{PARM+Show-o}, just naively depict a cat and a bed as distinct, unrelated objects. Similarly, in tasks involving logical operations or quantifiers such as the prompt ``\textit{either a spoon or a bowl}'' (Figure~\ref{fig:appendix_logical}), most models incorrectly render both objects, reflecting an inability to correctly interpret disjunctive semantics. 

We hypothesize that these limitations are rooted in the bag-of-words encoding mechanism used by CLIP-based conditioning in diffusion models. A formal investigation of this hypothesis is left for future work.

\vspace{-2mm}
\paragraph{Mathematical Reasoning Remains a Significant Bottleneck.} Across all reasoning categories, T2I models exhibit profound limitations in addressing mathematical reasoning tasks. Most models achieve near-zero accuracy on this front. Notably, even the best-performing open-source model, \modelname{SD3-medium}, attains a score of merely $0.19$, while others, including \modelname{LlamaGen}, \modelname{Show-o}, and \modelname{ELLA$_\text{SD1.5}$}, score below $0.10$.
As shown in Figure~\ref{fig:appendix_mathematical}, prompts involving geometric transformations (e.g., ``\textit{rotate a square 90°}'') frequently result in irrelevant outputs such as abstract art or clocks. Similarly, prompts grounded in number theory (e.g., ``\textit{visualize the twin prime pairs below 50}'') yield outputs like mecha robots (\modelname{EMU3}) or glowing, non-descriptive artifacts (\modelname{Show-o+PARM}, \modelname{Show-o}). These observations indicate a severe lack of training data containing mathematical visual concepts, hindering the models’ ability to perform reliable numerical or mathematical reasoning.

\begin{figure*}[htbp]

\centering
\includegraphics[width=0.9\textwidth]{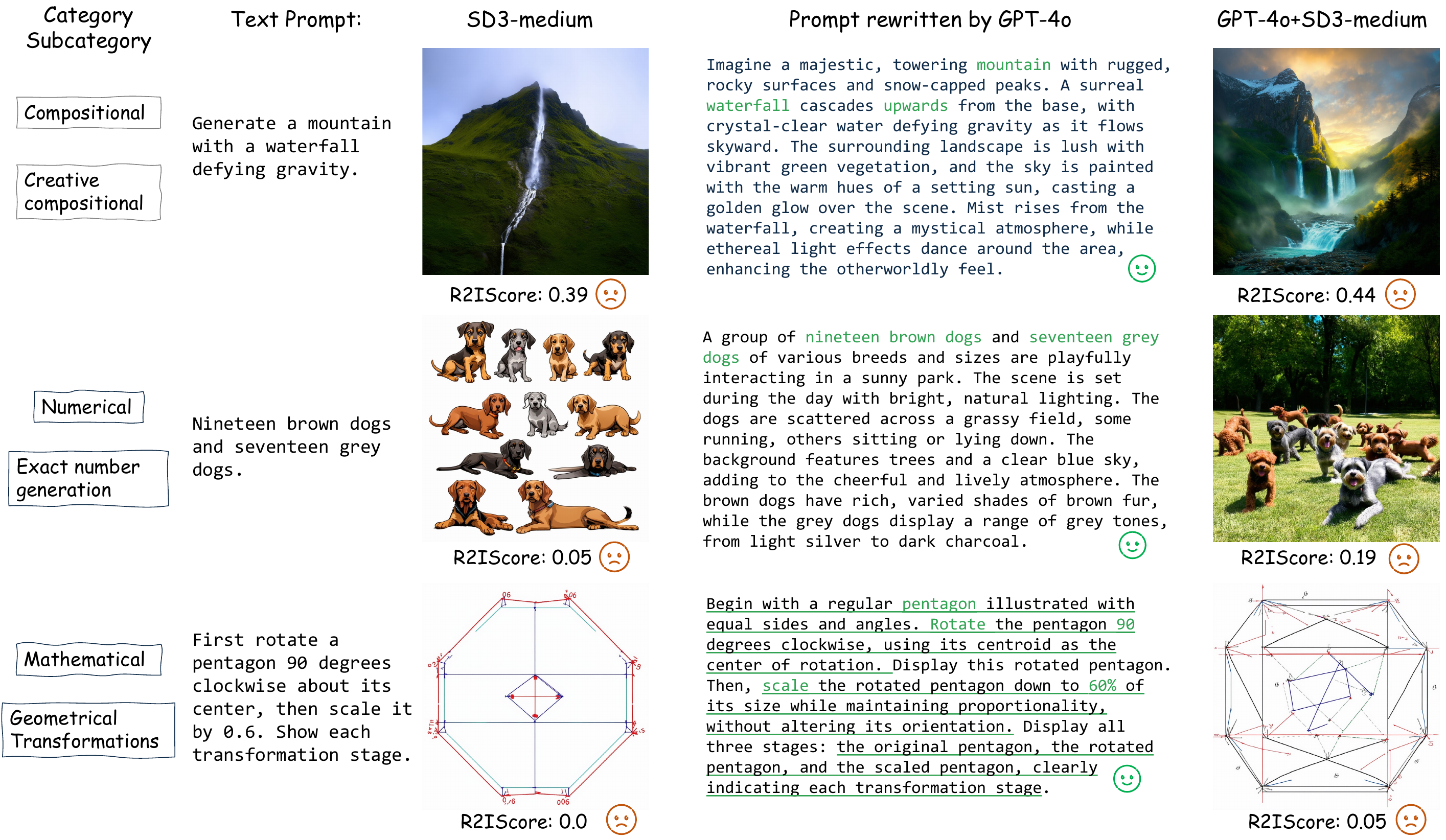}

\caption{Failure Cases of the Pipeline-based Framework on Compositional/Numerical/Mathematical Reasoning.}
\label{fig:llm_failure}

\end{figure*}

\vspace{-2mm}
\paragraph{Marginal Improvements from Reasoning-Enhanced Architectures.} Reasoning-enhanced models such as \modelname{Show-o+PARM}, \modelname{Show-o+ORM}, and \modelname{Show-o+DPO} demonstrate only incremental improvements over their respective base models. For example, the best-performing variant (i.e., \modelname{Show-o+PARM}) achieves an overall score of $0.38$, compared to $0.36$ achieved by the base model \modelname{Show-o}. Notably, these models continue to perform poorly on the most challenging categories, including mathematical reasoning ($\leq 0.13$) and causal reasoning ($\leq 0.32$), 
indicating that current methods, such as PARM (Potential Assessment Reward Model), ORM (Outcome Reward Model), and DPO (Direct Preference Optimization), offer limited improvements in reasoning-driven T2I generation. These results highlight the urgent need for more effective and targeted approaches for reasoning-driven T2I generation.

\vspace{-2mm}
\paragraph{Closed-Source Models Set the Upper Bound for Current Reasoning Capabilities.} Proprietary models such as \modelname{DALLE-3} and \modelname{gpt-image-1} significantly outperform their open-source counterparts, achieving $57.8\%$ and $71.1\%$ higher score than the best-performing open-source model (i.e., \modelname{SD3-medium}), respectively. Notably, \modelname{gpt-image-1} consistently achieves the highest scores across all reasoning categories. This significant performance gap highlights the pressing need for open, reproducible benchmarks and the development of competitive open-source T2I models to bridge the capability gap with proprietary systems.

\vspace{-2mm}
\paragraph{Pipeline-based T2I Framework Improves Commonsense, Causal Reasoning, but Yields Marginal Gains for Compositional, Numerical and Mathematical Reasoning.}
As shown in Table~\ref{tab:main_table}, the pipeline-based framework yields substantial gains in all reasoning categories by an average of $0.13$, e.g., improvements ranging from $0.21$ to $0.58$ are observed in causal reasoning, commonsense reasoning. A detailed comparison across fine-grained reasoning subcategories is shown in Figure~\ref{fig:llm-rewrite-overall-performance}. Despite the general effectiveness of the pipeline-based framework, gains in \textit{Compositional}, \textit{Mathematical} categories remain modest $(\leq 0.13)$. 
As shown in Figure~\ref{fig:llm_failure}, many reasoning concepts remain challenging for T2I models to faithfully render, even when clearly articulated by the LLM. In \textit{Compositional reasoning} (Example 1), despite GPT-4o correctly reasons that ``\textit{a surreal waterfall cascades upwards from the base},'' \modelname{SD3-medium} still renders a downward-flowing waterfall. In \textit{Numerical reasoning} (Example 2), although \modelname{GPT-4o} expands the original prompt ``\textit{Nineteen brown dogs and seventeen grey dogs}'' with additional detail, the generated image fails to depict the correct number of dogs. For \textit{Mathematical reasoning} (Example 3), the difficulty goes beyond language to abstract cognition: although GPT-4o specifies terms like ``\textit{display all three stages}'' and ``\textit{regular pentagon},'' the output remains visually inaccurate, with \modelname{SD3-medium} producing disorganized geometric shapes.

Success in this domain often requires models to grasp geometric structures such as points, lines, angles, and spatial transformations. We posit that overcoming these limitations will require not only more mathematically enriched training data but also the integration of architectural components or external modules capable of reasoning over structured symbolic knowledge.

\subsection{Evaluation of \metric}\label{sec:comparisionevaluation}

We further assess the effectiveness of our proposed \metric by evaluating its alignment with human judgments. We conduct a human study involving a group of senior college students, where each participant compares the image outputs generated by two T2I models, Lumina-Image 2.0~\cite{qin2025lumina} and Sana-1.5~\cite{xie2025sana15efficientscaling}, and selects the image that best aligns with the prompts or indicates if both are equally satisfactory or unsatisfactory. 
More details are provided in Appendix~\ref{sec:humanannotators}. 
We also apply \metric to evaluate the same set of image pairs and compute its judgements with those of human annotators, 
using three established evaluation metrics: \textit{Pairwise Accuracy}~\cite{deutsch2023ties}, \textit{Kendall's $\tau$}~\cite{jadhav2019kendall}, and \textit{Spearman Correlation}~\cite{tu2025between}. We compare \metric against several widely adopted T2I generation evaluation metrics, including \textbf{DSGscore}~\cite{cho2023davidsonian}, \textbf{VIEScore}~\cite{ku2023viescore}, \textbf{CLIPScore}~\cite{hessel2021clipscore}, and \textbf{VQA score}~\cite{lin2024evaluating}.
Since these existing metrics mainly focus on surface-level text-image alignment and image quality, \metric consistently achieves superior alignment with human judgements across all alignment criteria, as shown in Table~\ref{tab:evaluationcomparision}, 
demonstrating its effectiveness and robustness as an evaluation metric of reasoning-driven T2I generation. Further experimental details and additional results are provided in Appendix~\ref{sec:appendix_metric}.

\renewcommand{\arraystretch}{1.5} 
\begin{table}[htbp]
\centering
\footnotesize

\resizebox{\columnwidth}{!}{
\begin{tabular}{lcccc}
\toprule
\textbf{Models} & \textbf{Pairwise Accuracy} & \textbf{Kendall's $\tau$} & \textbf{Spearman Correlation}  \\

\midrule
CLIPScore & 0.631 & 0.263 & 0.310  \\ 
DSGScore & 0.520 & 0.220 & 0.254  \\
VIEScore & 0.694 & 0.494 & 0.451  \\
VQAscore & 0.629 & 0.463 & 0.563  \\
\midrule
\metric  & \textbf{0.713} & \textbf{0.747} & \textbf{0.694} \\
\bottomrule
\end{tabular}}

\caption{Comparison of \metric with other Evaluation Metrics for T2I Generation.}

\label{tab:evaluationcomparision}
\end{table}

\begin{figure}[htbp]
	\centering
	\includegraphics[width=\linewidth ]{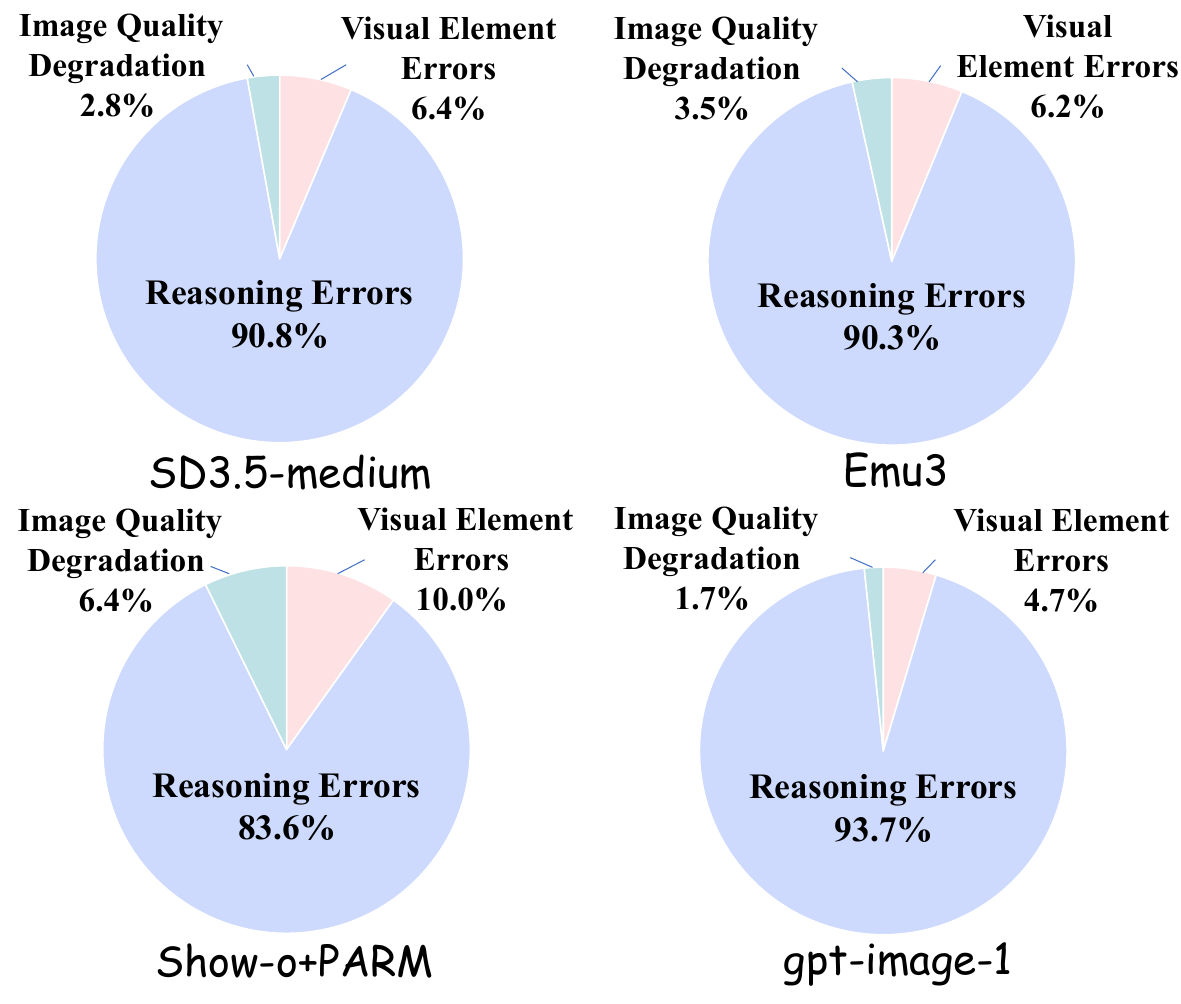}
    \vspace{-5mm}
	\caption{Distribution of Errors of \modelname{Emu3}, \modelname{SD3-medium}, \modelname{Show-o+PARM}, \modelname{gpt-image-1}.}
    \label{fig:erroranalysis}
 \vspace{-5mm}
\end{figure}

\subsection{Error Analysis} \label{sec:error-analysis}

To better understand the limitations of current T2I models, we categorize 
and accordingly define three failure types: \textit{basic element errors}, \textit{reasoning errors}, and \textit{visual quality issues}. For qualitative analysis, we examine representative models from each architectural category, including \modelname{Emu3}, \modelname{SD3-medium}, \modelname{Show-o+PARM}, and \modelname{gpt-image-1}. The relative distribution of these failure types is computed and visualized in Figure~\ref{fig:erroranalysis}. As we can see, reasoning-related failures dominate the error distribution across all four models, accounting for over 80\% of total errors. This observation highlights reasoning as the primary bottleneck in current T2I systems. Among the evaluated models, \modelname{Show-o+PARM} exhibits a relatively higher proportion of basic element errors, suggesting its limitation in accurately rendering basic visual components. In contrast, \modelname{gpt-image-1} demonstrates the lowest rates of both basic element and image quality errors, indicating its superior performance in both semantic fidelity and visual rendering.

\section{Conclusion}

This paper introduces \benchmark, a comprehensive benchmark designed to evaluate the reasoning capabilities of text-to-image (T2I) generation models across $7$ core reasoning categories and $32$ subcategories. Alongside \benchmark, we design \metric, a QA-style evaluation metric specifically tailored for reasoning-driven T2I generation, with stronger correlation with human judgments compared to existing evaluation metrics.
Our evaluation reveals consistently limited reasoning capabilities across all existing T2I models, highlighting the pressing need for more robust, reasoning-aware T2I generation architectures.

\section*{Limitations}

\paragraph{Evaluation Method Constraints} Despite our diligent efforts to design and refine evaluation questions and criteria for each data instance, aimed at enhancing reasoning-based evaluation, the current method is inherently constrained by the specific benchmark used in this study. As such, it cannot be directly generalized to other datasets without further adaptations. Although the manually crafted evaluation questions and criteria facilitate the use of vision language models for scoring, leading to more transparent and interpretable evaluations, the granularity of these evaluations remains relatively coarse compared to the detailed assessments conducted at the training level. Future work could focus on the development of a versatile reward model tailored for evaluating Text-to-Image (T2I) reasoning generation, which would also support reinforcement learning from Human Feedback (RLHF).

\paragraph{Language and Dataset Scope} At present, our evaluation of T2I models is confined to \benchmark, which is based solely on English-language data. Consequently, the reasoning capabilities of models in non-English language contexts remain unexplored. Additionally, some models do not support symbolic inputs, such as emojis or complex mathematical notations. For the sake of ensuring the benchmark’s general applicability, we have excluded data instances that feature such symbolic inputs. Besides, our benchmark is limited only to image generation. Extending to video/audio/3D generation can be another promising future direction.

\section*{Ethics Statement}
Some instances in our dataset were generated using GPT-4o, a powerful language model that has been designed to simulate human-like text generation. Although this model produces high-quality outputs, it is important to note that the generated content reflects the biases and limitations inherent in the training data. We are aware of the ethical implications of using such models, especially in terms of the potential for reinforcing harmful stereotypes or generating inappropriate content. In this study, we have made efforts to mitigate these risks by carefully curating the dataset and implementing a manual review process. However, we acknowledge that there may still be residual biases present, and we encourage future work to focus on developing methods to reduce such biases, ensuring that generated content aligns with ethical guidelines and societal norms.

\bibliography{acl_latex}

\begin{thebibliography}{51}
\expandafter\ifx\csname natexlab\endcsname\relax\def\natexlab#1{#1}\fi

\bibitem[{Chen et~al.(2024)Chen, Tan, Cai, Xie, Zhao, Zhang, Lin, Bai, Liu, and Chang}]{chen2024spark}
Liang Chen, Sinan Tan, Zefan Cai, Weichu Xie, Haozhe Zhao, Yichi Zhang, Junyang Lin, Jinze Bai, Tianyu Liu, and Baobao Chang. 2024.
\newblock A spark of vision-language intelligence: 2-dimensional autoregressive transformer for efficient finegrained image generation.
\newblock In \emph{The Thirteenth International Conference on Learning Representations}.

\bibitem[{Chen et~al.(2025)Chen, Wu, Liu, Pan, Liu, Xie, Yu, and Ruan}]{chen2025janus}
Xiaokang Chen, Zhiyu Wu, Xingchao Liu, Zizheng Pan, Wen Liu, Zhenda Xie, Xingkai Yu, and Chong Ruan. 2025.
\newblock Janus-pro: Unified multimodal understanding and generation with data and model scaling.
\newblock \emph{arXiv preprint arXiv:2501.17811}.

\bibitem[{Chevalley et~al.(2022)Chevalley, Roohani, Mehrjou, Leskovec, and Schwab}]{chevalley2022causalbench}
Mathieu Chevalley, Yusuf Roohani, Arash Mehrjou, Jure Leskovec, and Patrick Schwab. 2022.
\newblock Causalbench: A large-scale benchmark for network inference from single-cell perturbation data.
\newblock \emph{arXiv preprint arXiv:2210.17283}.

\bibitem[{Cho et~al.(2023)Cho, Hu, Garg, Anderson, Krishna, Baldridge, Bansal, Pont-Tuset, and Wang}]{cho2023davidsonian}
Jaemin Cho, Yushi Hu, Roopal Garg, Peter Anderson, Ranjay Krishna, Jason Baldridge, Mohit Bansal, Jordi Pont-Tuset, and Su~Wang. 2023.
\newblock Davidsonian scene graph: Improving reliability in fine-grained evaluation for text-to-image generation.
\newblock \emph{arXiv preprint arXiv:2310.18235}.

\bibitem[{Deutsch et~al.(2023)Deutsch, Foster, and Freitag}]{deutsch2023ties}
Daniel Deutsch, George Foster, and Markus Freitag. 2023.
\newblock Ties matter: Meta-evaluating modern metrics with pairwise accuracy and tie calibration.
\newblock \emph{arXiv preprint arXiv:2305.14324}.

\bibitem[{Esser et~al.(2024{\natexlab{a}})Esser, Kulal, Blattmann, Entezari, M{\"u}ller, Saini, Levi, Lorenz, Sauer, Boesel et~al.}]{esser2024scaling}
Patrick Esser, Sumith Kulal, Andreas Blattmann, Rahim Entezari, Jonas M{\"u}ller, Harry Saini, Yam Levi, Dominik Lorenz, Axel Sauer, Frederic Boesel, et~al. 2024{\natexlab{a}}.
\newblock Scaling rectified flow transformers for high-resolution image synthesis.
\newblock In \emph{Forty-first international conference on machine learning}.

\bibitem[{Esser et~al.(2024{\natexlab{b}})Esser, Kulal, Blattmann, Entezari, Müller, Saini, Levi, Lorenz, Sauer, Boesel, Podell, Dockhorn, English, Lacey, Goodwin, Marek, and Rombach}]{esser2024scalingrectifiedflowtransformers}
Patrick Esser, Sumith Kulal, Andreas Blattmann, Rahim Entezari, Jonas Müller, Harry Saini, Yam Levi, Dominik Lorenz, Axel Sauer, Frederic Boesel, Dustin Podell, Tim Dockhorn, Zion English, Kyle Lacey, Alex Goodwin, Yannik Marek, and Robin Rombach. 2024{\natexlab{b}}.
\newblock \href {http://arxiv.org/abs/2403.03206} {Scaling rectified flow transformers for high-resolution image synthesis}.

\bibitem[{Fu et~al.(2024)Fu, He, Lu, Wang, and Roth}]{fu2024commonsense}
Xingyu Fu, Muyu He, Yujie Lu, William~Yang Wang, and Dan Roth. 2024.
\newblock Commonsense-t2i challenge: Can text-to-image generation models understand commonsense?
\newblock \emph{arXiv preprint arXiv:2406.07546}.

\bibitem[{Gao et~al.(2024)Gao, Zhuo, Liu, Du, Luo, Qiu, Zhang, Lin, Huang, Geng et~al.}]{gao2024lumina}
Peng Gao, Le~Zhuo, Dongyang Liu, Ruoyi Du, Xu~Luo, Longtian Qiu, Yuhang Zhang, Chen Lin, Rongjie Huang, Shijie Geng, et~al. 2024.
\newblock Lumina-t2x: Transforming text into any modality, resolution, and duration via flow-based large diffusion transformers.
\newblock \emph{arXiv preprint arXiv:2405.05945}.

\bibitem[{Ghosh et~al.(2023)Ghosh, Hajishirzi, and Schmidt}]{ghosh2023geneval}
Dhruba Ghosh, Hannaneh Hajishirzi, and Ludwig Schmidt. 2023.
\newblock Geneval: An object-focused framework for evaluating text-to-image alignment.
\newblock \emph{Advances in Neural Information Processing Systems}, 36:52132--52152.

\bibitem[{Guo et~al.(2025)Guo, Zhang, Tong, Zhao, Gao, Li, and Heng}]{guo2025can}
Ziyu Guo, Renrui Zhang, Chengzhuo Tong, Zhizheng Zhao, Peng Gao, Hongsheng Li, and Pheng-Ann Heng. 2025.
\newblock Can we generate images with cot? let's verify and reinforce image generation step by step.
\newblock \emph{arXiv preprint arXiv:2501.13926}.

\bibitem[{Hessel et~al.(2021)Hessel, Holtzman, Forbes, Bras, and Choi}]{hessel2021clipscore}
Jack Hessel, Ari Holtzman, Maxwell Forbes, Ronan~Le Bras, and Yejin Choi. 2021.
\newblock Clipscore: A reference-free evaluation metric for image captioning.
\newblock \emph{arXiv preprint arXiv:2104.08718}.

\bibitem[{Hu et~al.(2024)Hu, Wang, Fang, Fu, Cheng, and Yu}]{hu2024ella}
Xiwei Hu, Rui Wang, Yixiao Fang, Bin Fu, Pei Cheng, and Gang Yu. 2024.
\newblock Ella: Equip diffusion models with llm for enhanced semantic alignment.
\newblock \emph{arXiv preprint arXiv:2403.05135}.

\bibitem[{Hurst et~al.(2024)Hurst, Lerer, Goucher, Perelman, Ramesh, Clark, Ostrow, Welihinda, Hayes, Radford et~al.}]{hurst2024gpt}
Aaron Hurst, Adam Lerer, Adam~P Goucher, Adam Perelman, Aditya Ramesh, Aidan Clark, AJ~Ostrow, Akila Welihinda, Alan Hayes, Alec Radford, et~al. 2024.
\newblock Gpt-4o system card.
\newblock \emph{arXiv preprint arXiv:2410.21276}.

\bibitem[{Jadhav and Ma(2019)}]{jadhav2019kendall}
Sneha Jadhav and Shuangge Ma. 2019.
\newblock Kendall's tau for functional data analysis.
\newblock \emph{arXiv preprint arXiv:1912.03725}.

\bibitem[{Jiang et~al.(2025)Jiang, Guo, Zhang, Zong, Li, Zhuo, Yan, Heng, and Li}]{jiang2025t2i}
Dongzhi Jiang, Ziyu Guo, Renrui Zhang, Zhuofan Zong, Hao Li, Le~Zhuo, Shilin Yan, Pheng-Ann Heng, and Hongsheng Li. 2025.
\newblock T2i-r1: Reinforcing image generation with collaborative semantic-level and token-level cot.
\newblock \emph{arXiv preprint arXiv:2505.00703}.

\bibitem[{Kaji{\'c} et~al.(2024)Kaji{\'c}, Wiles, Albuquerque, Bauer, Wang, Pont-Tuset, and Nematzadeh}]{kajic2024evaluating}
Ivana Kaji{\'c}, Olivia Wiles, Isabela Albuquerque, Matthias Bauer, Su~Wang, Jordi Pont-Tuset, and Aida Nematzadeh. 2024.
\newblock Evaluating numerical reasoning in text-to-image models.
\newblock \emph{Advances in Neural Information Processing Systems}, 37:42211--42224.

\bibitem[{Ku et~al.(2023)Ku, Jiang, Wei, Yue, and Chen}]{ku2023viescore}
Max Ku, Dongfu Jiang, Cong Wei, Xiang Yue, and Wenhu Chen. 2023.
\newblock Viescore: Towards explainable metrics for conditional image synthesis evaluation.
\newblock \emph{arXiv preprint arXiv:2312.14867}.

\bibitem[{Ku et~al.(2024)Ku, Jiang, Wei, Yue, and Chen}]{ku2024viescoreexplainablemetricsconditional}
Max Ku, Dongfu Jiang, Cong Wei, Xiang Yue, and Wenhu Chen. 2024.
\newblock \href {http://arxiv.org/abs/2312.14867} {Viescore: Towards explainable metrics for conditional image synthesis evaluation}.

\bibitem[{Lee et~al.(2023)Lee, Yasunaga, Meng, Mai, Park, Gupta, Zhang, Narayanan, Teufel, Bellagente et~al.}]{lee2023holistic}
Tony Lee, Michihiro Yasunaga, Chenlin Meng, Yifan Mai, Joon~Sung Park, Agrim Gupta, Yunzhi Zhang, Deepak Narayanan, Hannah Teufel, Marco Bellagente, et~al. 2023.
\newblock Holistic evaluation of text-to-image models.
\newblock \emph{Advances in Neural Information Processing Systems}, 36:69981--70011.

\bibitem[{Lei et~al.(2025)Lei, Zhang, Hu, Lin, Li, Sun, Du, Zhuo, Li, Li et~al.}]{lei2025imagine}
Jiayi Lei, Renrui Zhang, Xiangfei Hu, Weifeng Lin, Zhen Li, Wenjian Sun, Ruoyi Du, Le~Zhuo, Zhongyu Li, Xinyue Li, et~al. 2025.
\newblock Imagine-e: Image generation intelligence evaluation of state-of-the-art text-to-image models.
\newblock \emph{arXiv preprint arXiv:2501.13920}.

\bibitem[{Li et~al.(2024)Li, Lin, Pathak, Li, Fei, Wu, Ling, Xia, Zhang, Neubig et~al.}]{li2024genai}
Baiqi Li, Zhiqiu Lin, Deepak Pathak, Jiayao Li, Yixin Fei, Kewen Wu, Tiffany Ling, Xide Xia, Pengchuan Zhang, Graham Neubig, et~al. 2024.
\newblock Genai-bench: Evaluating and improving compositional text-to-visual generation.
\newblock \emph{arXiv preprint arXiv:2406.13743}.

\bibitem[{Liao et~al.(2025)Liao, Yang, Li, Li, Lin, Cheng, and Wang}]{liao2025imagegen}
Jiaqi Liao, Zhengyuan Yang, Linjie Li, Dianqi Li, Kevin Lin, Yu~Cheng, and Lijuan Wang. 2025.
\newblock Imagegen-cot: Enhancing text-to-image in-context learning with chain-of-thought reasoning.
\newblock \emph{arXiv preprint arXiv:2503.19312}.

\bibitem[{Liew et~al.(2022)Liew, Yan, Zhou, and Feng}]{liew2022magicmix}
Jun~Hao Liew, Hanshu Yan, Daquan Zhou, and Jiashi Feng. 2022.
\newblock Magicmix: Semantic mixing with diffusion models.
\newblock \emph{arXiv preprint arXiv:2210.16056}.

\bibitem[{Lin et~al.(2024)Lin, Pathak, Li, Li, Xia, Neubig, Zhang, and Ramanan}]{lin2024evaluating}
Zhiqiu Lin, Deepak Pathak, Baiqi Li, Jiayao Li, Xide Xia, Graham Neubig, Pengchuan Zhang, and Deva Ramanan. 2024.
\newblock Evaluating text-to-visual generation with image-to-text generation.
\newblock In \emph{European Conference on Computer Vision}, pages 366--384. Springer.

\bibitem[{Liu et~al.(2020)Liu, Cui, Liu, Huang, Wang, and Zhang}]{liu2020logiqa}
Jian Liu, Leyang Cui, Hanmeng Liu, Dandan Huang, Yile Wang, and Yue Zhang. 2020.
\newblock Logiqa: A challenge dataset for machine reading comprehension with logical reasoning.
\newblock \emph{arXiv preprint arXiv:2007.08124}.

\bibitem[{Liu et~al.(2025)Liu, Ma, Yang, Dan, Yu, Zhao, Hu, Liu, and Fan}]{liu2025llm4gen}
Mushui Liu, Yuhang Ma, Zhen Yang, Jun Dan, Yunlong Yu, Zeng Zhao, Zhipeng Hu, Bai Liu, and Changjie Fan. 2025.
\newblock Llm4gen: Leveraging semantic representation of llms for text-to-image generation.
\newblock In \emph{Proceedings of the AAAI Conference on Artificial Intelligence}, volume~39, pages 5523--5531.

\bibitem[{Lu et~al.(2023)Lu, Yang, Li, Wang, and Wang}]{lu2023llmscore}
Yujie Lu, Xianjun Yang, Xiujun Li, Xin~Eric Wang, and William~Yang Wang. 2023.
\newblock Llmscore: Unveiling the power of large language models in text-to-image synthesis evaluation.
\newblock \emph{Advances in Neural Information Processing Systems}, 36:23075--23093.

\bibitem[{Ma et~al.(2024)Ma, Zheng, Wei, Wu, Lu, Zhang, Xie, Gong, Zhu, and Shen}]{ma2024learning}
Shuailei Ma, Kecheng Zheng, Ying Wei, Wei Wu, Fan Lu, Yifei Zhang, Chen-Wei Xie, Biao Gong, Jiapeng Zhu, and Yujun Shen. 2024.
\newblock Learning visual generative priors without text.
\newblock \emph{arXiv preprint arXiv:2412.07767}.

\bibitem[{Marino et~al.(2019)Marino, Rastegari, Farhadi, and Mottaghi}]{marino2019ok}
Kenneth Marino, Mohammad Rastegari, Ali Farhadi, and Roozbeh Mottaghi. 2019.
\newblock Ok-vqa: A visual question answering benchmark requiring external knowledge.
\newblock In \emph{Proceedings of the IEEE/cvf conference on computer vision and pattern recognition}, pages 3195--3204.

\bibitem[{Niu et~al.(2025)Niu, Ning, Zheng, Lin, Jin, Liao, Ning, Zhu, and Yuan}]{niu2025wise}
Yuwei Niu, Munan Ning, Mengren Zheng, Bin Lin, Peng Jin, Jiaqi Liao, Kunpeng Ning, Bin Zhu, and Li~Yuan. 2025.
\newblock Wise: A world knowledge-informed semantic evaluation for text-to-image generation.
\newblock \emph{arXiv preprint arXiv:2503.07265}.

\bibitem[{Qin et~al.(2025{\natexlab{a}})Qin, Zhuo, Xin, Du, Li, Fu, Lu, Li, Liu, Zhu, Beddow, Millon, Victor~Perez, Qiao, Zhang, Liu, Li, Xu, and Gao}]{lumina2}
Qi~Qin, Le~Zhuo, Yi~Xin, Ruoyi Du, Zhen Li, Bin Fu, Yiting Lu, Xinyue Li, Dongyang Liu, Xiangyang Zhu, Will Beddow, Erwann Millon, Wenhai~Wang Victor~Perez, Yu~Qiao, Bo~Zhang, Xiaohong Liu, Hongsheng Li, Chang Xu, and Peng Gao. 2025{\natexlab{a}}.
\newblock \href {http://arxiv.org/abs/2503.21758} {Lumina-image 2.0: A unified and efficient image generative framework}.

\bibitem[{Qin et~al.(2025{\natexlab{b}})Qin, Zhuo, Xin, Du, Li, Fu, Lu, Yuan, Li, Liu et~al.}]{qin2025lumina}
Qi~Qin, Le~Zhuo, Yi~Xin, Ruoyi Du, Zhen Li, Bin Fu, Yiting Lu, Jiakang Yuan, Xinyue Li, Dongyang Liu, et~al. 2025{\natexlab{b}}.
\newblock Lumina-image 2.0: A unified and efficient image generative framework.
\newblock \emph{arXiv preprint arXiv:2503.21758}.

\bibitem[{Rombach et~al.(2022)Rombach, Blattmann, Lorenz, Esser, and Ommer}]{rombach2022high}
Robin Rombach, Andreas Blattmann, Dominik Lorenz, Patrick Esser, and Bjorn Ommer. 2022.
\newblock High-resolution image synthesis with latent diffusion models.
\newblock \emph{Proceedings of the IEEE/CVF Conference on Computer Vision and Pattern Recognition (CVPR)}, pages 10684--10695.

\bibitem[{Sun et~al.(2024{\natexlab{a}})Sun, Jiang, Chen, Zhang, Peng, Luo, and Yuan}]{sun2024autoregressive}
Peize Sun, Yi~Jiang, Shoufa Chen, Shilong Zhang, Bingyue Peng, Ping Luo, and Zehuan Yuan. 2024{\natexlab{a}}.
\newblock Autoregressive model beats diffusion: Llama for scalable image generation.
\newblock \emph{arXiv preprint arXiv:2406.06525}.

\bibitem[{Sun et~al.(2024{\natexlab{b}})Sun, Cui, Zhang, Zhang, Yu, Luo, Wang, Rao, Liu, Huang, and Wang}]{sun2024generativemultimodalmodelsincontext}
Quan Sun, Yufeng Cui, Xiaosong Zhang, Fan Zhang, Qiying Yu, Zhengxiong Luo, Yueze Wang, Yongming Rao, Jingjing Liu, Tiejun Huang, and Xinlong Wang. 2024{\natexlab{b}}.
\newblock \href {http://arxiv.org/abs/2312.13286} {Generative multimodal models are in-context learners}.

\bibitem[{Sun et~al.(2023)Sun, Yu, Cui, Zhang, Zhang, Wang, Gao, Liu, Huang, and Wang}]{sun2023emu}
Quan Sun, Qiying Yu, Yufeng Cui, Fan Zhang, Xiaosong Zhang, Yueze Wang, Hongcheng Gao, Jingjing Liu, Tiejun Huang, and Xinlong Wang. 2023.
\newblock Emu: Generative pretraining in multimodality.
\newblock \emph{arXiv preprint arXiv:2307.05222}.

\bibitem[{Thrush et~al.(2022)Thrush, Jiang, Bartolo, Singh, Williams, Kiela, and Ross}]{thrush2022winoground}
Tristan Thrush, Ryan Jiang, Max Bartolo, Amanpreet Singh, Adina Williams, Douwe Kiela, and Candace Ross. 2022.
\newblock Winoground: Probing vision and language models for visio-linguistic compositionality.
\newblock In \emph{Proceedings of the IEEE/CVF Conference on Computer Vision and Pattern Recognition}, pages 5238--5248.

\bibitem[{Tong et~al.(2024)Tong, Fan, Zhu, Xiong, Chen, Sinha, Rabbat, LeCun, Xie, and Liu}]{tong2024metamorph}
Shengbang Tong, David Fan, Jiachen Zhu, Yunyang Xiong, Xinlei Chen, Koustuv Sinha, Michael Rabbat, Yann LeCun, Saining Xie, and Zhuang Liu. 2024.
\newblock Metamorph: Multimodal understanding and generation via instruction tuning.
\newblock \emph{arXiv preprint arXiv:2412.14164}.

\bibitem[{Tu et~al.(2025)Tu, Li, and Shepherd}]{tu2025between}
Shengxin Tu, Chun Li, and Bryan~E Shepherd. 2025.
\newblock Between-and within-cluster spearman rank correlations.
\newblock \emph{Statistics in Medicine}, 44(3-4):e10326.

\bibitem[{Wang et~al.(2024)Wang, Zhang, Luo, Sun, Cui, Wang, Zhang, Wang, Li, Yu et~al.}]{wang2024emu3}
Xinlong Wang, Xiaosong Zhang, Zhengxiong Luo, Quan Sun, Yufeng Cui, Jinsheng Wang, Fan Zhang, Yueze Wang, Zhen Li, Qiying Yu, et~al. 2024.
\newblock Emu3: Next-token prediction is all you need.
\newblock \emph{arXiv preprint arXiv:2409.18869}.

\bibitem[{Wu et~al.(2024)Wu, Yu, Huang, Russakovsky, and Arora}]{wu2024conceptmix}
Xindi Wu, Dingli Yu, Yangsibo Huang, Olga Russakovsky, and Sanjeev Arora. 2024.
\newblock Conceptmix: A compositional image generation benchmark with controllable difficulty.
\newblock \emph{arXiv preprint arXiv:2408.14339}.

\bibitem[{Xiao et~al.(2024)Xiao, Wang, Zhou, Yuan, Xing, Yan, Li, Wang, Huang, and Liu}]{xiao2024omnigen}
Shitao Xiao, Yueze Wang, Junjie Zhou, Huaying Yuan, Xingrun Xing, Ruiran Yan, Chaofan Li, Shuting Wang, Tiejun Huang, and Zheng Liu. 2024.
\newblock Omnigen: Unified image generation.
\newblock \emph{arXiv preprint arXiv:2409.11340}.

\bibitem[{Xie et~al.(2025{\natexlab{a}})Xie, Chen, Zhao, Yu, Zhu, Wu, Lin, Zhang, Li, Chen, Cai, Liu, Zhou, and Han}]{xie2025sana15efficientscaling}
Enze Xie, Junsong Chen, Yuyang Zhao, Jincheng Yu, Ligeng Zhu, Chengyue Wu, Yujun Lin, Zhekai Zhang, Muyang Li, Junyu Chen, Han Cai, Bingchen Liu, Daquan Zhou, and Song Han. 2025{\natexlab{a}}.
\newblock \href {http://arxiv.org/abs/2501.18427} {Sana 1.5: Efficient scaling of training-time and inference-time compute in linear diffusion transformer}.

\bibitem[{Xie et~al.(2025{\natexlab{b}})Xie, Chen, Zhao, Yu, Zhu, Wu, Lin, Zhang, Li, Chen et~al.}]{xie2025sana}
Enze Xie, Junsong Chen, Yuyang Zhao, Jincheng Yu, Ligeng Zhu, Chengyue Wu, Yujun Lin, Zhekai Zhang, Muyang Li, Junyu Chen, et~al. 2025{\natexlab{b}}.
\newblock Sana 1.5: Efficient scaling of training-time and inference-time compute in linear diffusion transformer.
\newblock \emph{arXiv preprint arXiv:2501.18427}.

\bibitem[{Xie et~al.(2024)Xie, Mao, Bai, Zhang, Wang, Lin, Gu, Chen, Yang, and Shou}]{xie2024show}
Jinheng Xie, Weijia Mao, Zechen Bai, David~Junhao Zhang, Weihao Wang, Kevin~Qinghong Lin, Yuchao Gu, Zhijie Chen, Zhenheng Yang, and Mike~Zheng Shou. 2024.
\newblock Show-o: One single transformer to unify multimodal understanding and generation.
\newblock \emph{arXiv preprint arXiv:2408.12528}.

\bibitem[{Yang et~al.(2024)Yang, Liu, Hong, Zhang, Huang, Cai, Zhang, and Cui}]{yang2024improvingdiffusionbasedimagesynthesis}
Ling Yang, Jingwei Liu, Shenda Hong, Zhilong Zhang, Zhilin Huang, Zheming Cai, Wentao Zhang, and Bin Cui. 2024.
\newblock \href {http://arxiv.org/abs/2401.02015} {Improving diffusion-based image synthesis with context prediction}.

\bibitem[{Zhang et~al.(2024)Zhang, Dai, Yang, An, Feng, and Ren}]{zhang2024var}
Qian Zhang, Xiangzi Dai, Ninghua Yang, Xiang An, Ziyong Feng, and Xingyu Ren. 2024.
\newblock Var-clip: Text-to-image generator with visual auto-regressive modeling.
\newblock \emph{arXiv preprint arXiv:2408.01181}.

\bibitem[{Zhao et~al.(2025)Zhao, Zhang, Tang, Li, Zhang, Zhai, Yan, Yang, Yang, and Duan}]{zhao2025envisioning}
Xiangyu Zhao, Peiyuan Zhang, Kexian Tang, Hao Li, Zicheng Zhang, Guangtao Zhai, Junchi Yan, Hua Yang, Xue Yang, and Haodong Duan. 2025.
\newblock Envisioning beyond the pixels: Benchmarking reasoning-informed visual editing.
\newblock \emph{arXiv preprint arXiv:2504.02826}.

\bibitem[{Zhou et~al.(2024)Zhou, Yu, Babu, Tirumala, Yasunaga, Shamis, Kahn, Ma, Zettlemoyer, and Levy}]{zhou2024transfusion}
Chunting Zhou, Lili Yu, Arun Babu, Kushal Tirumala, Michihiro Yasunaga, Leonid Shamis, Jacob Kahn, Xuezhe Ma, Luke Zettlemoyer, and Omer Levy. 2024.
\newblock Transfusion: Predict the next token and diffuse images with one multi-modal model.
\newblock \emph{arXiv preprint arXiv:2408.11039}.

\bibitem[{Zhu et~al.(2024)Zhu, Sun, Song, Xiao, Li, Wang, Huang, Yang, and Xu}]{zhu2024evaluating}
Xiangru Zhu, Penglei Sun, Yaoxian Song, Yanghua Xiao, Zhixu Li, Chengyu Wang, Jun Huang, Bei Yang, and Xiaoxiao Xu. 2024.
\newblock Evaluating semantic variation in text-to-image synthesis: A causal perspective.
\newblock \emph{arXiv preprint arXiv:2410.10291}.

\end{thebibliography}
\appendix

\clearpage

\clearpage
\section{More Details about \benchmark}
\subsection{Statistics of \benchmark}
\label{sec:stat-of-benchmark}
We list the statistics of \benchmark in Table \ref{tab:key_statistics}.

\begin{table}[htbp]
    \small
\centering
\renewcommand{\arraystretch}{1}  
\begin{adjustbox}{width=\linewidth}
   \begin{tabular}{lr}
     \toprule
     \textbf{Statistic} & \textbf{Number} \\
     \midrule
     Total data instances & 3,068 \\
     ~- \cellcolor{commonsense_green} Commonsense & \cellcolor{commonsense_green}695 (22.65\%) \\
     ~- \cellcolor{composition_yellow} Compositional & \cellcolor{composition_yellow}311 (10.14\%) \\
     ~- \cellcolor{numerical_orange} Numerical & \cellcolor{numerical_orange}322 (10.50\%) \\
     ~- \cellcolor{causal_rose} Causal & \cellcolor{causal_rose}151 (4.92\%) \\
     ~- \cellcolor{mathematical_blue} Mathematical & \cellcolor{mathematical_blue}800 (26.08\%) \\
     ~- \cellcolor{logical_purple} Logical & \cellcolor{logical_purple} 630 (20.53\%) \\
     ~- \cellcolor{concept_mixing_red} Concept Mixing & \cellcolor{concept_mixing_red}159 (5.18\%) \\
     \midrule
     Categories & 7 \\
     Subcategories & 32 \\
     Evaluation dimensions & 3 \\
     \midrule
     Vocabulary size & 7,184 \\
     Maximum prompt length & 35 \\
     Maximum reference caption length & 28 \\
     Maximum evaluation questions & 18 \\
     Average prompt length & 21.7 \\
     Average reference caption length & 23.4 \\
     Average evaluation questions & 12.2 \\
     \bottomrule
   \end{tabular}
\end{adjustbox}
\caption{\textbf{Key Statistics of \benchmark.}}
\label{tab:key_statistics}
\end{table}

\subsection{Image by Categories}
\label{sec:visualization_subcategory}
This section presents examples of images from various categories in \benchmark. Figure~\ref{fig:appendix_commonsense} to~\ref{fig:appendix_compositional} coresponding to images under the categories of ~\textit{Commonsense Reasoning, Numerical Reasoning, Causal Reasoning, Logical Reasoning, Mathematical Reasoning, Concept Mixing Reasoning, Compositional Reasoning}, respectively. 

\subsection{Definition of Categories in \benchmark}
\label{sec:category_definitions}
\vspace{0.3cm}
The data instances in \benchmark encompass seven core categories: Commonsense Reasoning, Compositional Reasoning, Conceptual Mixing Reasoning, Numerical Reasoning, Logical Reasoning, Causal Reasoning, and Mathematical Reasoning. These categories are further subdivided into thirty-two more granular subcategories, providing a thorough evaluation of the reasoning capabilities of Text-to-Image (T2I) models.

\begin{figure*}
\centering
\includegraphics[width=\textwidth]{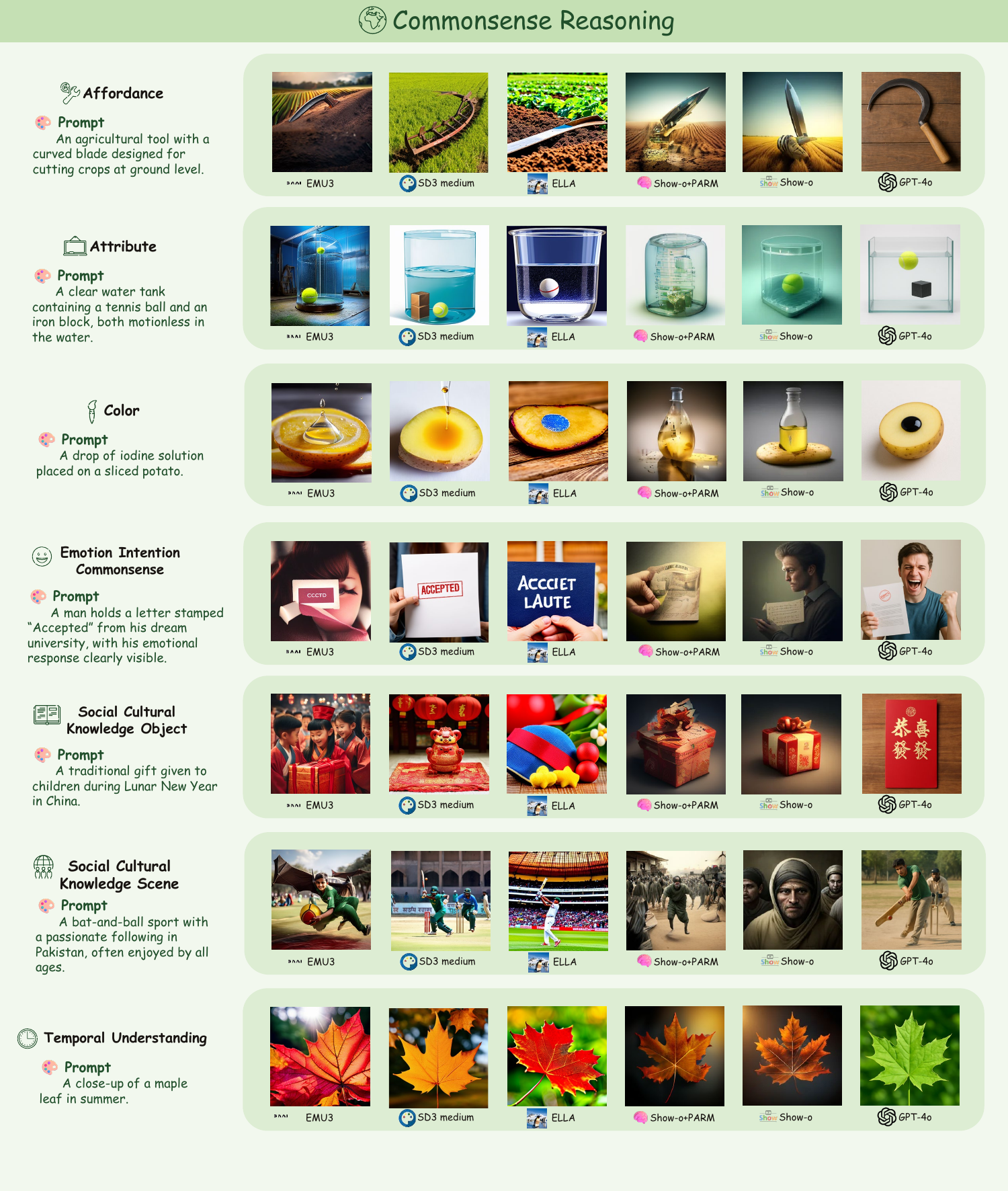}
   \caption{\textbf{Examples of Seven Subfields in Commonsense Reasoning}, spanning Affordance, Attribute, Color, Emotion Intention Commonsense, Social Cultural Knowledge Object and Scene and Temporal Understanding. We showcase the Text-lite version.}
\label{fig:appendix_commonsense}
\end{figure*}

\paragraph{\textbf{Commonsense Reasoning}}  
Commonsense reasoning is a critical aspect of evaluating a model's understanding of general knowledge and contextual information. It involves utilizing external \textit{knowledge resources}—such as \textit{world knowledge}, \textit{cultural context}, or \textit{background information}—to reason about the content of an image, rather than simply replicating the image. This allows for a richer context in assessing the commonsense reasoning capabilities of \textit{Text-to-Image} (T2I) models. In \benchmark, we categorize commonsense reasoning into seven distinct \textit{subfields}, as shown in Figure~\ref{fig:appendix_commonsense}, with detailed definitions provided in Table~\ref{tab:commosense_subcategory_definition}.

\begin{figure*}[htbp]
\centering
\includegraphics[width=\textwidth]{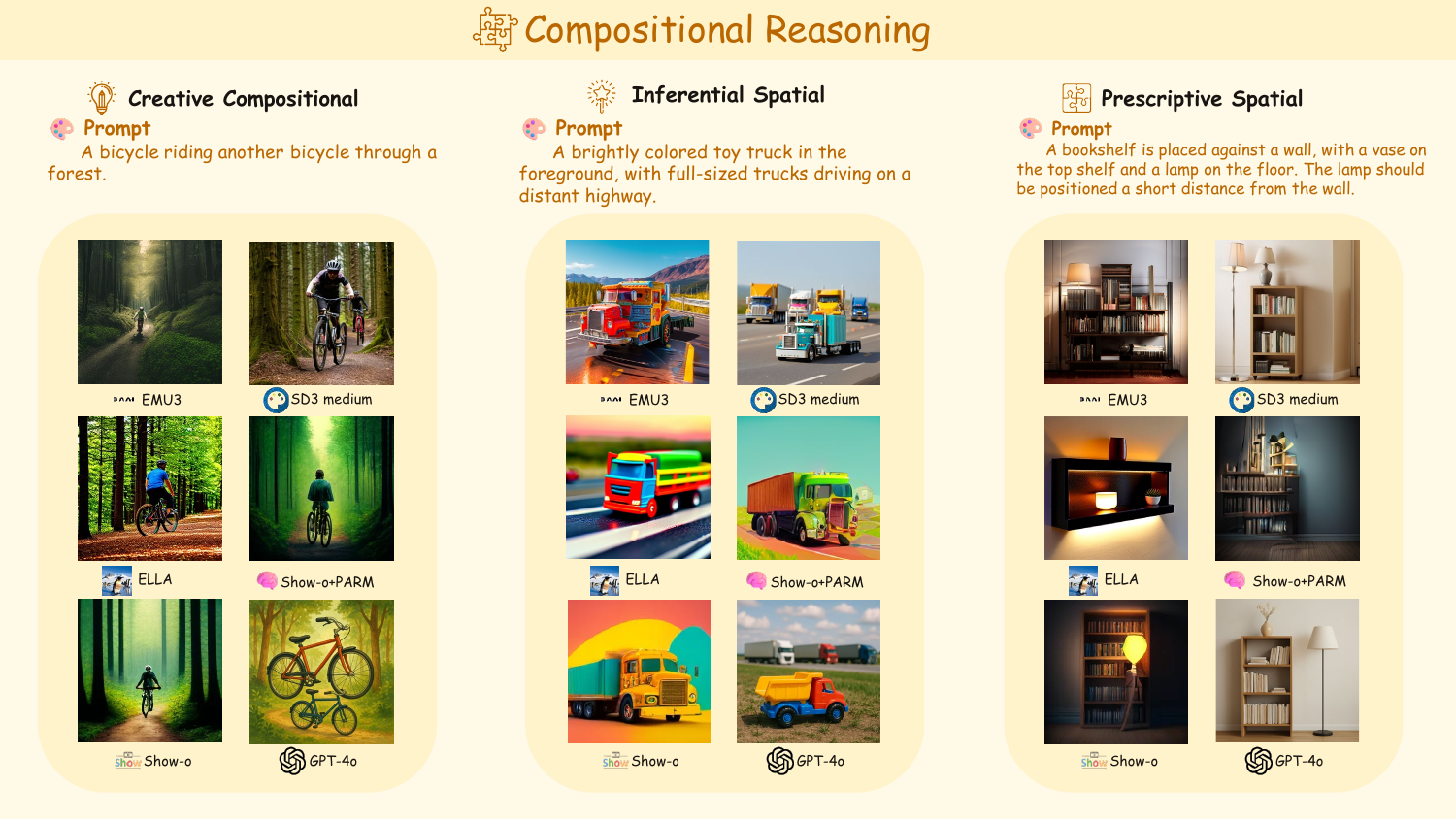}
   \caption{\textbf{Examples of Three Subfields in Compositional Reasoning}, including Creative Compositional, Inferential Spatial, Color, Prescriptive Spatial.}
\label{fig:appendix_compositional}

\end{figure*}

\paragraph{\textbf{Compositional Reasoning}}  
Compositional reasoning refers to the ability to combine smaller, simpler \textit{components} or pieces of \textit{information} to form more complex \textit{concepts}, \textit{solutions}, or \textit{conclusions}. It involves understanding the \textit{relationships} between individual parts and how they contribute to the whole, enabling \textit{logical reasoning} within structured, hierarchical, or layered systems. In \benchmark, we divide compositional reasoning into three \textit{subfields}, as depicted in Figure~\ref{fig:appendix_compositional}, with their definitions outlined in Table~\ref{tab:compositional_subcategory_definition}.

\begin{figure*}[htbp]
\centering
\includegraphics[width=\textwidth]{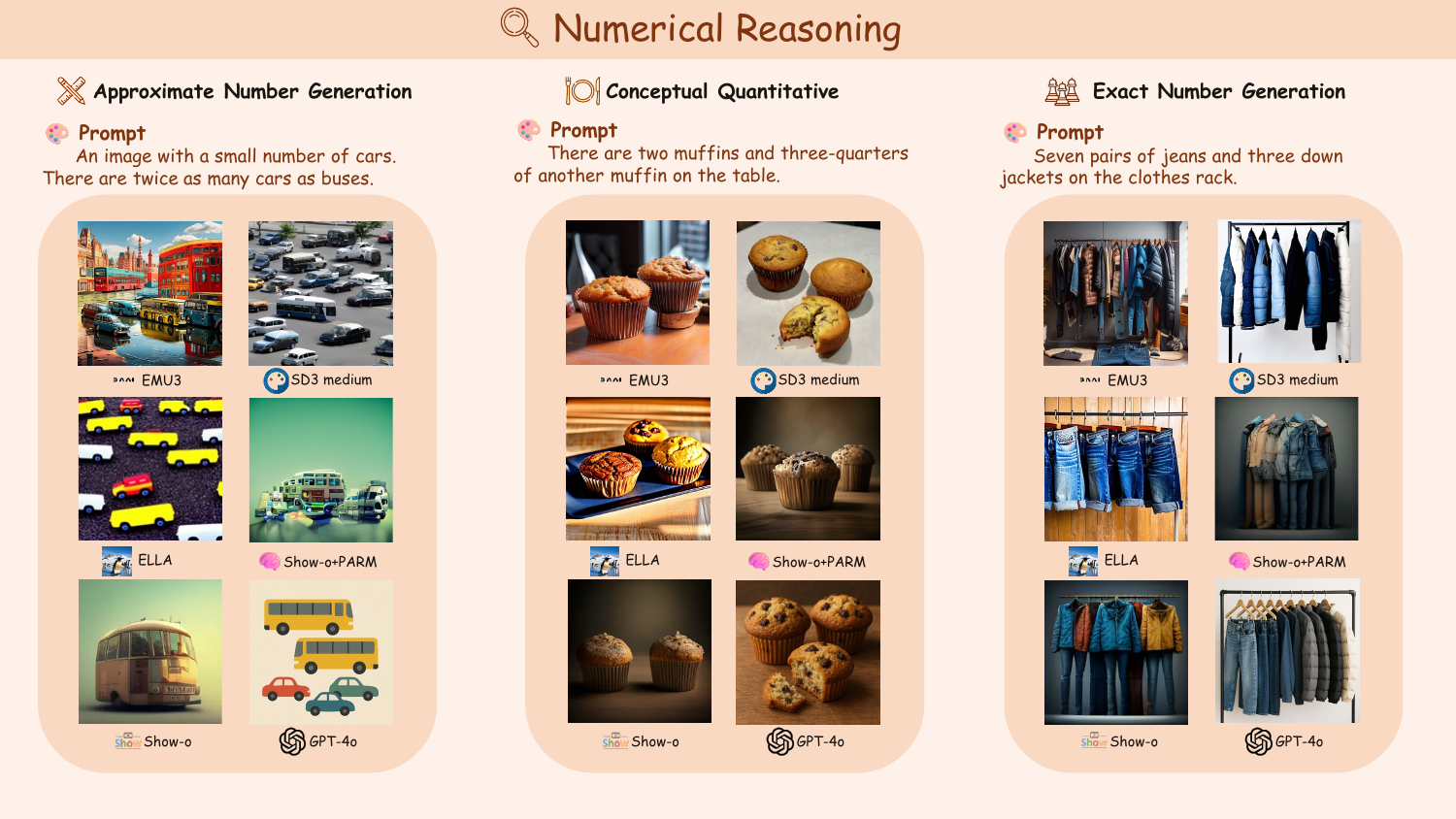}
   \caption{\textbf{Examples of Three Subfields in Numerical Reasoning}, including Approximate Number Generation, Conceptual Quantitative, Exact Number Generation.}
\label{fig:appendix_numerical}
\end{figure*}

\paragraph{\textbf{Numerical Reasoning}}  
Numerical reasoning, in the context of T2I models, involves the ability of these models to accurately interpret, process, and generate \textit{images} based on \textit{numerical information} presented in \textit{textual prompts}. In \benchmark, we categorize numerical reasoning into three \textit{subfields}, as illustrated in Figure~\ref{fig:appendix_numerical}, with definitions provided in Table~\ref{tab:numerical_subcategory_definition}.

\begin{figure*}[htbp]
\centering
\includegraphics[width=\textwidth]{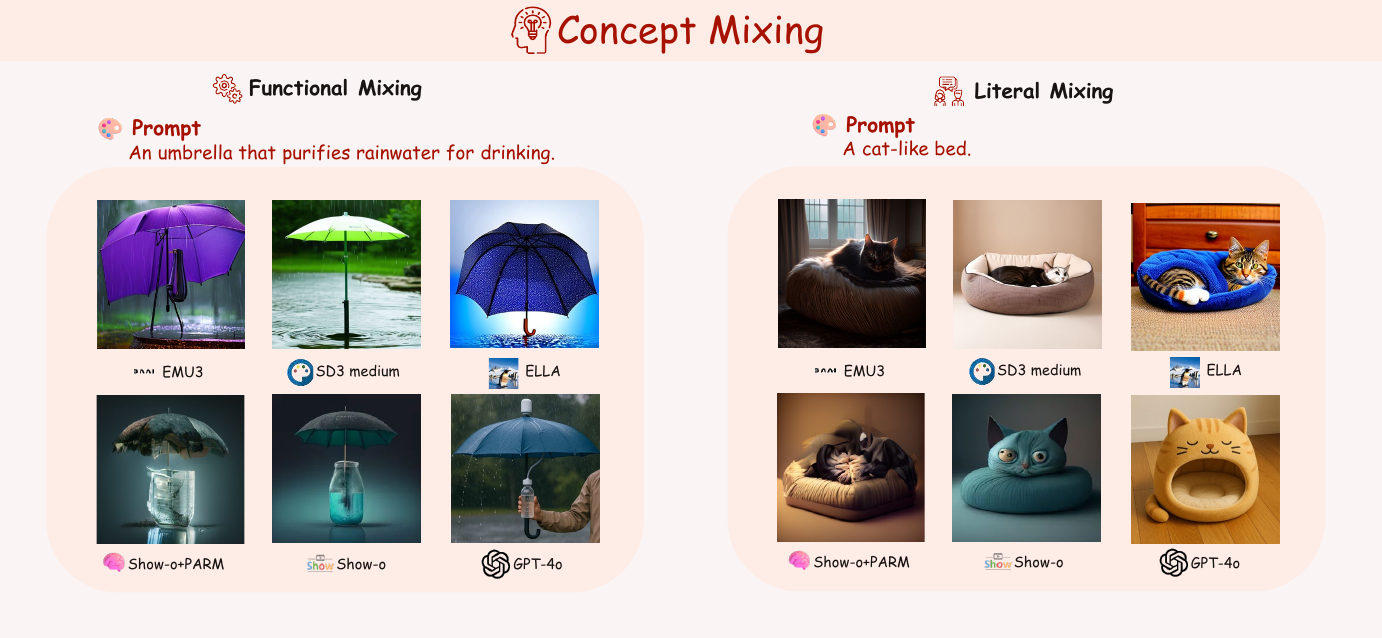}
   \caption{\textbf{Examples of Two Subfields in Concept Mixing}, including Functional Mixing and Literal Mixing. }
\label{fig:appendix_concept_mixing}

\end{figure*}

\paragraph{\textbf{Concept Mixing Reasoning}}  
Concept-Mixing reasoning refers to the process of combining different \textit{semantic elements} to create a new, unique \textit{concept}. In \benchmark, we divide concept-mixing reasoning into three \textit{subfields}, as shown in Figure~\ref{fig:appendix_concept_mixing}, with their definitions in Table~\ref{tab:concept_mixing_subcategory_definition}.

\begin{figure*}[htbp]
\centering
\includegraphics[width=\textwidth]{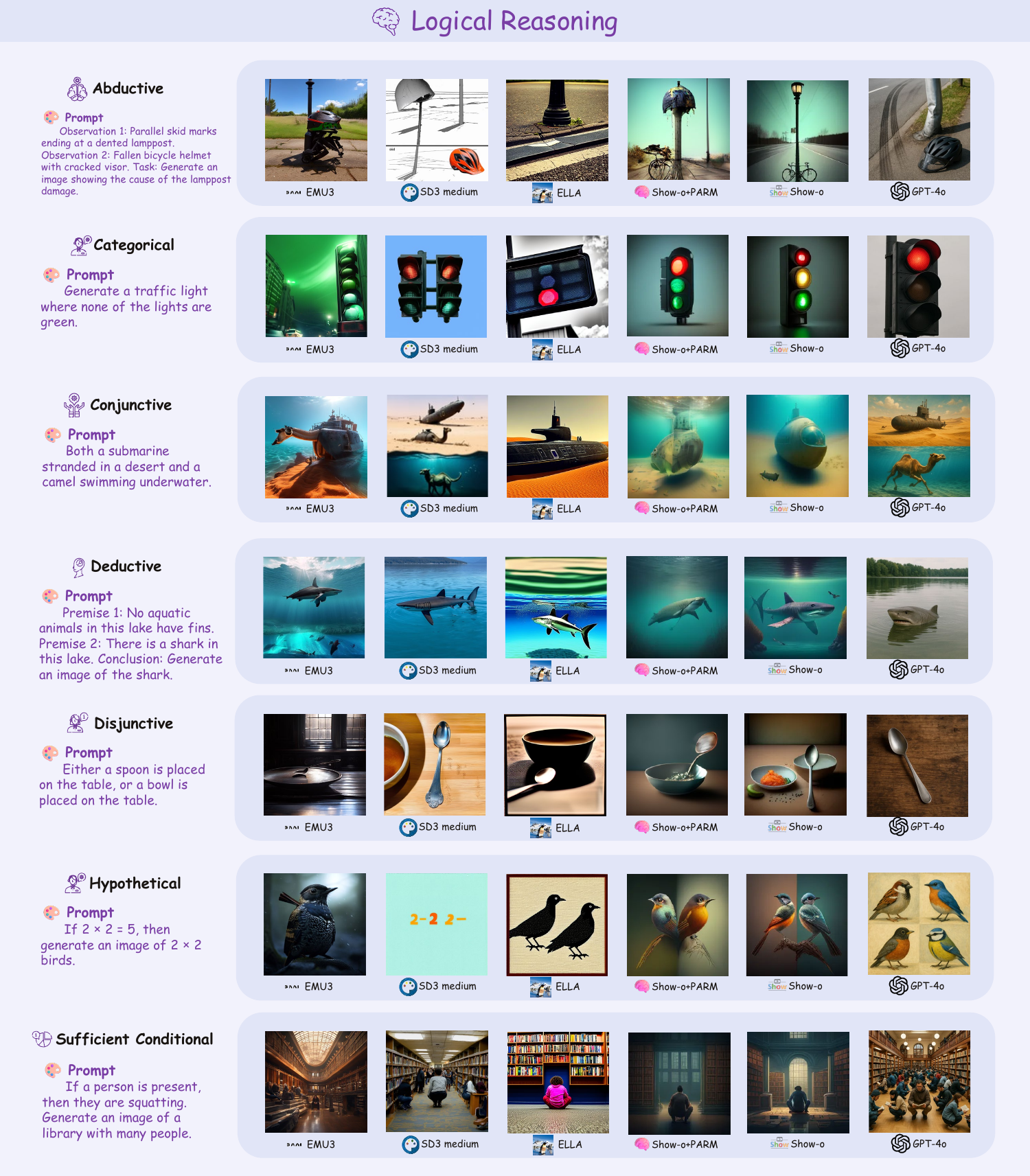}
   \caption{\textbf{Examples of Seven Subfields in Logical Reasoning}, spanning Abductive, Categorical, conjunctive, Deductive, Hypothetical, Sufficient Conditional. }
\label{fig:appendix_logical}

\end{figure*}

\paragraph{\textbf{Logical Reasoning}}  
Logical reasoning involves using systematic, structured approaches to analyze \textit{information}, draw \textit{conclusions}, and solve \textit{problems} based on given \textit{premises} or \textit{conditions}. In \benchmark, we break logical reasoning down into seven \textit{subfields}, as illustrated in Figure~\ref{fig:appendix_logical}, with definitions provided in Table~\ref{tab:logical_subcategory_definition}.

\begin{figure*}[htbp]
\centering
\includegraphics[width=\textwidth]{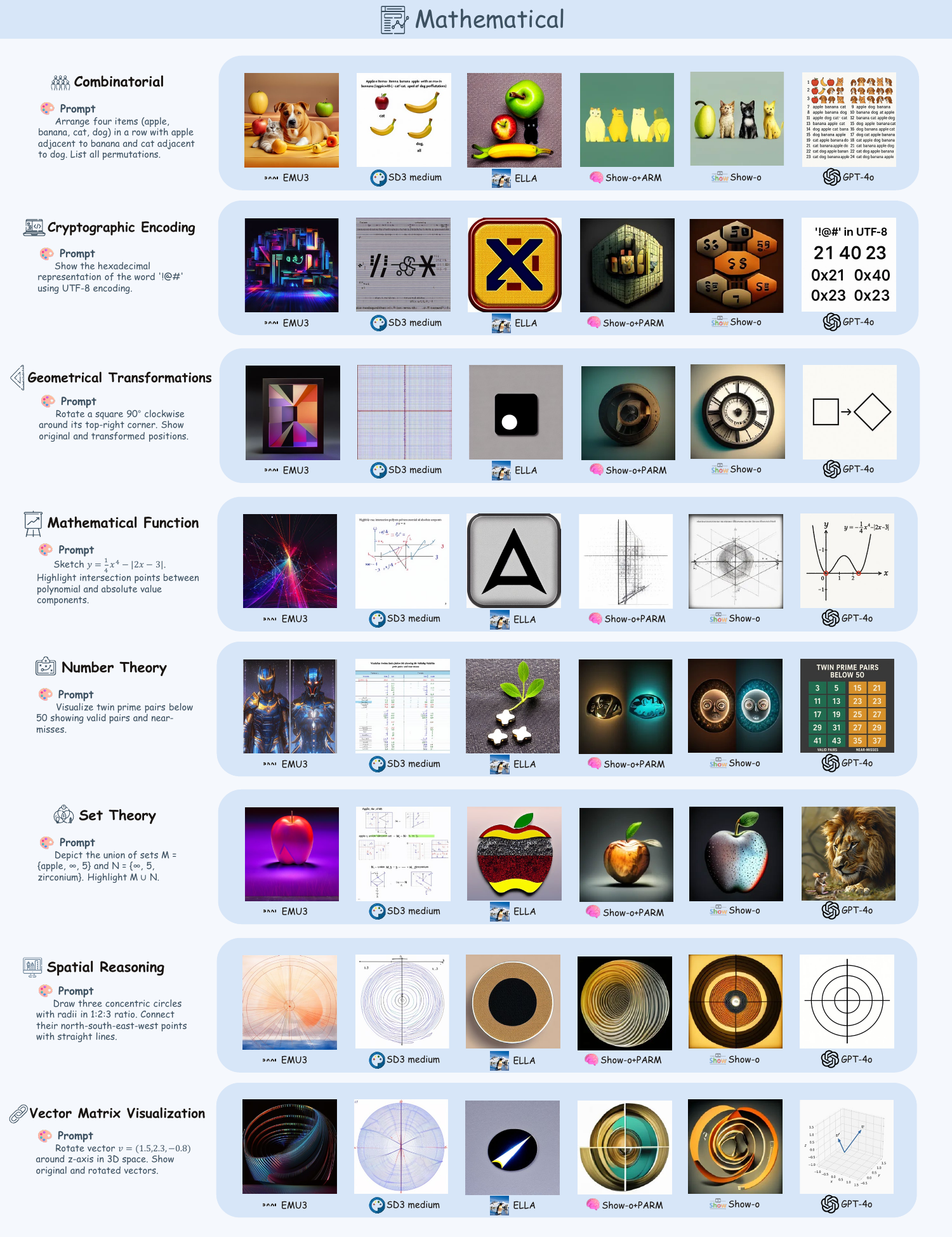}
   \caption{\textbf{Examples of Eight Subfields in Mathematical Reasoning}, spanning Combinatorial, Crypographic Encoding, Geometrical Transformations, Mathematical Function,spatial reasoning,et Theory, Spatial Reasoning and Vector Matrix Visualizations. }
\label{fig:appendix_mathematical}

\end{figure*}

\paragraph{\textbf{Mathematical Reasoning}}  
Mathematical reasoning refers to the ability to represent, understand, and generate visual representations of abstract \textit{mathematical concepts} and \textit{symbols}. In \benchmark, we subdivide mathematical reasoning into eight \textit{subfields}, as shown in Figure~\ref{fig:appendix_mathematical}, with their definitions outlined in Table~\ref{tab:mathematical_subcategory_definition}.

\paragraph{\textbf{Causal Reasoning}}  

\begin{figure*}[htbp]
\centering
\includegraphics[width=\textwidth]{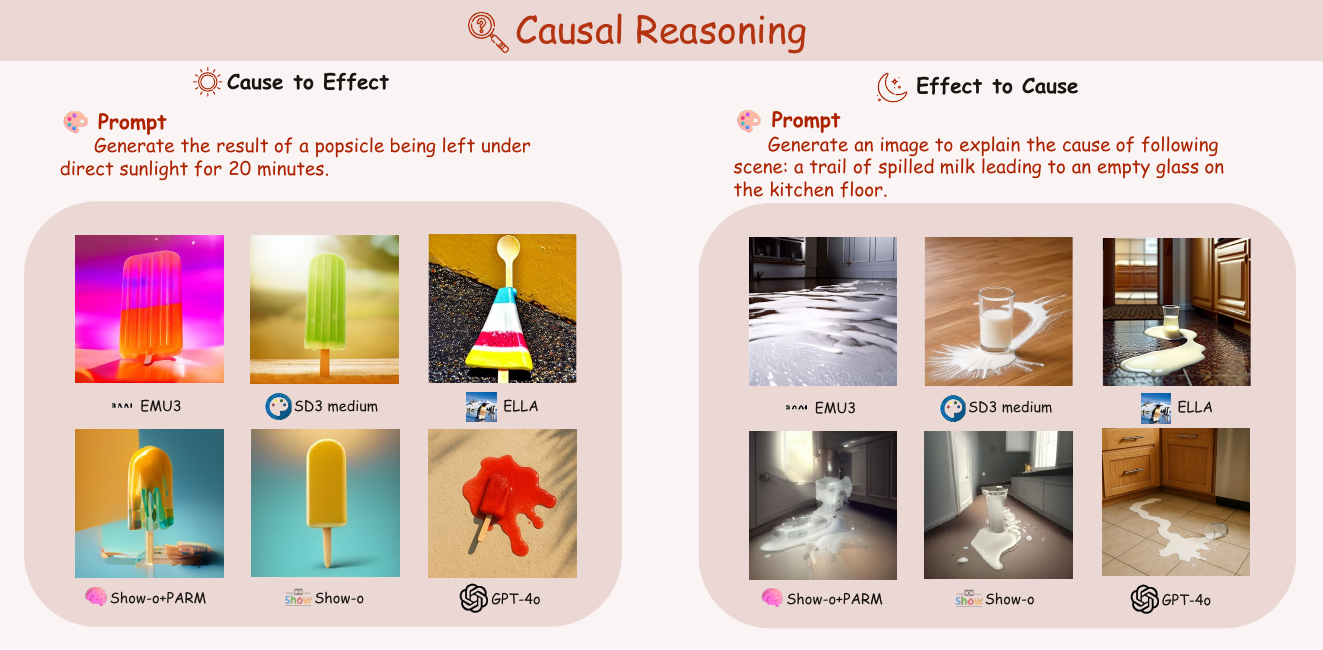}
   \caption{\textbf{Examples of two Subfields in Causal Reasoning}, including Cause to Effect Reasoning and Cause to Effect Reasoning. }
\label{fig:appendix_causal}

\end{figure*}

Causal reasoning is the ability to understand and explain \textit{cause-and-effect relationships}. In \benchmark, we categorize causal reasoning into three \textit{subfields}, as illustrated in Figure~\ref{fig:appendix_causal}, with definitions provided in Table~\ref{tab:causal_subcategory_definition}.

\subsection{Definition of Subcategories in \benchmark}
This section presents definitions of various subcategories under categories in \benchmark. Table ~\ref{tab:concept_mixing_subcategory_definition} to ~\ref{tab:mathematical_subcategory_definition} coresponding to subcategories under the categories of ~\textit{Commonsense Reasoning, Numerical Reasoning, Causal Reasoning, Logical Reasoning, Mathematical Reasoning, Concept Mixing Reasoning, Compositional Reasoning}, respectively.

\subsection{Data Generation Pipeline}
\label{sec:data-generation-pipeline}
\begin{figure*}[htbp]
\centering
\includegraphics[width=\textwidth]{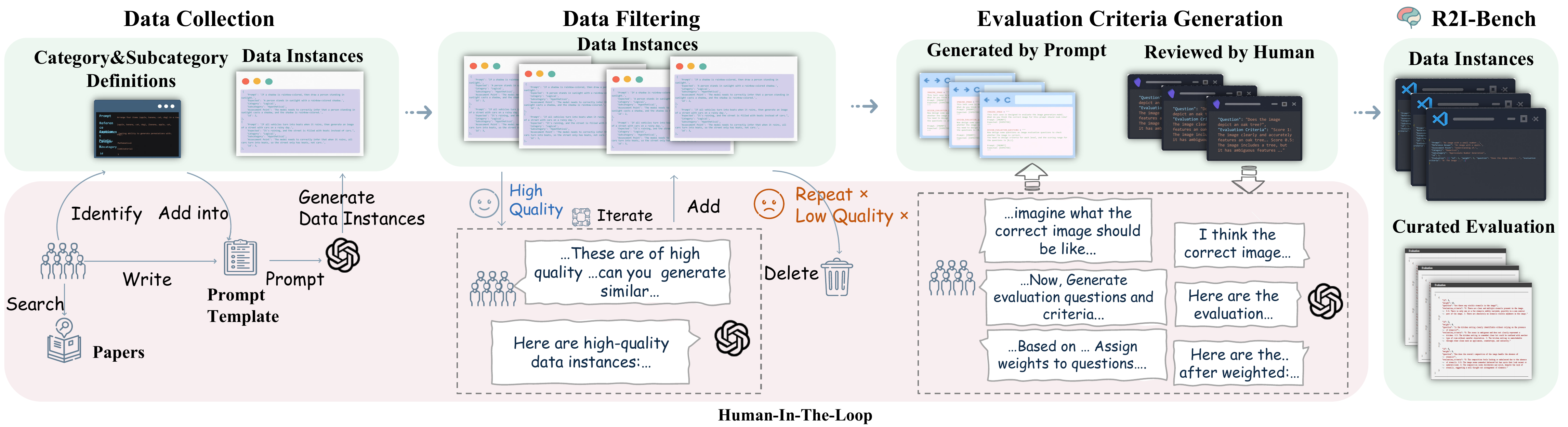}
\caption{\raggedright \textbf{Benchmark Curation Pipeline.} The pipeline starts with data collection, followed by data filtering, evaluation criteria generation, and ultimately results in \benchmark. To ensure data quality, human verification is performed at each key stage to eliminate low-quality data, annotations, and ambiguous evaluation questions.}
\label{fig:dataconstruction}
\vspace{-5mm}  
\end{figure*}

We build a human-in-the-loop data generation pipeline as illustrated in Figure~\ref{fig:dataconstruction}.

\begin{figure}[htbp]
\vspace{-15mm}  
\centering
\includegraphics[width=0.7\linewidth ]{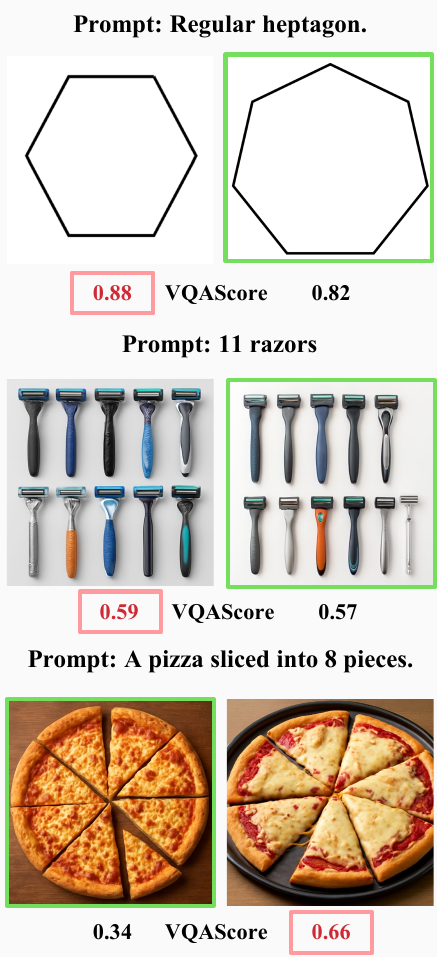}
\caption{\textbf{Failure Cases of VQAScore.}}
\label{fig:vqa_fail}
 \vspace{-3mm}
\end{figure}

\begin{table*}[htbp]
\centering
\small
\renewcommand\tabcolsep{2.5pt}  

\resizebox{\textwidth}{!}{\begin{tabular}{@{}l p{13cm}@{}}
    \toprule
    \textbf{Mathematical Reasoning} & \textbf{Description} \\
    \midrule
    
    \multirow{2.5}{*}{\parbox{4cm}{\centering Mathematical Function Visualization \\ (\text{12.50\%})}} 
    & Mathematical Function Visualization involves generating clear and informative images that depict \textit{mathematical functions}, their \textit{properties}, and the \textit{relationships} between various \textit{mathematical entities}, such as \textit{variables} and \textit{parameters}.    \\

    \cmidrule{1-2}
    
    \multirow{2}{*}{\parbox{4cm}{\centering Vector \& Matrix Visualization \\ (\text{12.50\%})}} 
    & Vector \& Matrix Visualization involves understanding and illustrating \textit{vectors}, \textit{matrices}, and \textit{transformations} in \textit{geometrical space}. \\
    
     \cmidrule{1-2}

    \multirow{2}{*}{\parbox{4cm}{\centering Combinatorial Reasoning \\ (\text{12.50\%})}} 
    & Combinatorial Reasoning involves depicting \textit{permutations}, \textit{combinations}, or \textit{arrangements} of \textit{objects}, often within a \textit{geometric} or \textit{graphical context}.
    \\   
 
    \cmidrule{1-2}
    
    \multirow{2}{*}{\parbox{4cm}{\centering Set Theory \& Relations \\ (\text{12.50\%})}} 
    & Set Theory \& Relations involves representing \textit{sets}, \textit{subsets}, and their \textit{relations} in visual forms (e.g., using \textit{Venn diagrams} or \textit{set-builder notation}).\\

     \cmidrule{1-2}
     
    \multirow{3}{*}{\parbox{4cm}{\centering Cryptographic \& Encoding Reasoning \\ (\text{12.50\%})}} 
    & Cryptographic \& Encoding Reasoning involves rendering \textit{encrypted texts}, \textit{ciphers}, or \textit{encoding schemes} (e.g., \textit{Morse code}, \textit{binary representations}).\\
      & \\ 
    \cmidrule{1-2}
    
    \multirow{2}{*}{\parbox{4cm}{\centering Number Theory \\ (\text{12.50\%})}} 
    & Number Theory Visualization involves depicting \textit{prime numbers}, \textit{divisibility rules}, and other abstract \textit{mathematical concepts}.\\

     \cmidrule{1-2}
     
    \multirow{2}{*}{\parbox{4cm}{\centering Geometrical Transformations \\ (\text{12.50\%})}} 
    &Geometrical Transformations involves illustrating \textit{symmetry operations} like \textit{rotations}, \textit{reflections}, \textit{translations}, or \textit{dilations} in space. \\
 
    \cmidrule{1-2}

    \multirow{2}{*}{\parbox{4cm}{\centering Spatial Reasoning \\ (\text{12.50\%})}} 
    &Spatial Reasoning refers to the ability to reason and infer the correct \textit{geometric configuration} of \textit{objects}, such as \textit{lines} and \textit{shapes}, in a defined space, based on specified \textit{spatial relationships}. \\

    \bottomrule
\end{tabular}}
\vspace{1ex}
\caption{Definitions and proportions of the eight subcategories in mathematical reasoning within \benchmark. The percentage indicates the proportion of each subcategory within the overall mathematical category.}
\label{tab:mathematical_subcategory_definition}
\end{table*}

\begin{table*}[htbp]
\centering
\small
\renewcommand\tabcolsep{2.5pt}
\resizebox{\textwidth}{!}{\begin{tabular}{@{}l p{12cm}@{}}
    \toprule
    \textbf{Concept Mixing Reasoning} & \textbf{Description} \\
    \midrule
    
    \multirow{2}{*}{\parbox{3cm}{\centering Functional Mixing \\ (\text{44.44\%})}} 
    & Functional mixing includes creating new \textit{concepts} that involve blending different \textit{functional properties} of \textit{objects}.    \\[2ex]  
    
    \cmidrule{1-2}
    
    \multirow{2}{*}{\parbox{3cm}{\centering Literal Mixing \\ (\text{55.56\%})}} 
    & Literal Mixing Reasoning combines \textit{elements} from different \textit{concepts} in a \textit{straightforward}, \textit{literal} manner, such as merging \textit{objects} or \textit{creatures}. \\

    \bottomrule
\end{tabular}}
\vspace{1ex}
\caption{Definitions and proportions of the two subcategories in concept mixing reasoning within \benchmark. The percentage indicates the proportion of each subcategory within the overall concept mixing category.}
\label{tab:concept_mixing_subcategory_definition}
\end{table*}

\begin{table*}[htbp]
\centering
\small
\renewcommand\tabcolsep{2.5pt}
\resizebox{\textwidth}{!}{\begin{tabular}{@{}l p{13cm}@{}}
    \toprule
    \textbf{Causal Reasoning} & \textbf{Description} \\
    \midrule
    
    \multirow{2}{*}{\parbox{3cm}{\centering Cause to Effect Reasoning \\ (\textit{52.98\%})}} 
    & Given a \textit{cause}, generate an \textit{image} depicting the \textit{effect}.    \\ 
    & \\  
    \cmidrule{1-2}
    
    \multirow{2}{*}{\parbox{3cm}{\centering Effect to Cause Reasoning \\ (\textit{47.02\%})}} 
    & Given an \textit{effect}, generate an \textit{image} depicting the possible \textit{cause}.\\ 
    & \\  

    \bottomrule
\end{tabular}}
\vspace{1ex}
\caption{Definitions and proportions of the two subcategories in causal reasoning within \benchmark. The percentage indicates the proportion of each subcategory within the overall causal category.}
\label{tab:causal_subcategory_definition}
\end{table*}

\begin{table*}[htbp]
\centering
\small
\renewcommand\tabcolsep{2.5pt}  

\resizebox{\textwidth}{!}{\begin{tabular}{@{}l p{13cm}@{}}
    \toprule
    \textbf{Compositional Reasoning} & \textbf{Description} \\
    \midrule
    \multirow{2.5}{*}{\parbox{4cm}{\centering Creative Composition Reasoning \\ (\text{32.15\%})}} 
    & Creative compositional reasoning is the ability to combine different \textit{ideas} or \textit{objects} in \textit{innovative} and \textit{imaginative} ways to create \textit{novel} and \textit{unique scenes} that have not been seen before. \\
    & \\ 

    \cmidrule{1-2}
    
    \multirow{3}{*}{\parbox{3.5cm}{\centering Inferential Spatial Reasoning \\ (\text{32.15\%})}} 
    & Inferential spatial reasoning refers to the ability to determine the \textit{positions} or \textit{size relationships} between \textit{objects} without explicit descriptions. \\
    & \\ 
    \cmidrule{1-2}
    
    \multirow{2.5}{*}{\parbox{3.5cm}{\centering Prescriptive Spatial Reasoning \\ (\text{35.69\%})}} 
    & Prescriptive Spatial Reasoning refers to the ability to follow clear \textit{instructions} about where \textit{objects} should be placed in a scene, ensuring the layout matches the described \textit{relationships}. Understanding phrases like \textit{"left of", "above", "inside"}. \\
    \bottomrule
\end{tabular}}
\vspace{1ex}
\caption{Definitions and proportions of the three subcategories in compositional reasoning within \benchmark. The percentage indicates the proportion of each subcategory within the overall compositional reasoning category.}
\label{tab:compositional_subcategory_definition}
\end{table*}
\begin{table*}[htbp]
\centering
\small
\renewcommand\tabcolsep{2.5pt}  

\resizebox{\textwidth}{!}{\begin{tabular}{@{}l p{13cm}@{}}
    \toprule
    \textbf{Logical Reasoning} & \textbf{Description} \\
    \midrule
    
    \multirow{2}{*}{\parbox{3cm}{\centering Categorical Reasoning \\ (\text{11.90\%})}} 
    & Categorical reasoning involves determining whether a specific \textit{concept} belongs to a particular \textit{category}. This type of reasoning often involves \textit{quantifiers} such as \textit{"all,"}, \textit{"everyone,"}, \textit{"any,"}, \textit{"no,"} and \textit{"some."}\\
    \cmidrule{1-2}
    
    \multirow{2}{*}{\parbox{3cm}{\centering Hypothetical Reasoning \\ (\text{11.90\%})}} 
    & Hypothetical reasoning is the process of using a \textit{systematic}, \textit{structured} approach to analyze \textit{information}, draw \textit{conclusions}, and solve \textit{problems} based on given \textit{premises} or \textit{conditions}. \\ 
      
    \cmidrule{1-2}
    
    \multirow{2}{*}{\parbox{3cm}{\centering Disjunctive Reasoning \\ (\text{16.51\%})}} 
    & Disjunctive reasoning involves \textit{premises} in the form \textit{"either ... or ..."}, where the \textit{conclusion} holds as long as one \textit{premise} is true.\\
      
    \cmidrule{1-2}

    \multirow{2}{*}{\parbox{3cm}{\centering Conjunctive Reasoning \\ (\text{16.51\%})}} 
    & Conjunctive reasoning involves \textit{premises} in the form \textit{"both ... and ..."}, where the \textit{conclusion} holds only if all the \textit{premises} is true.\\
  
    \cmidrule{1-2}

    \multirow{2.5}{*}{\parbox{3.2cm}{\centering Sufficient Conditional Reasoning \\ (\text{13.49\%})}} 
    & Sufficient Conditional Reasoning is based on \textit{conditional statements} of the form \textit{"If P, then Q"}, in which P is the \textit{antecedent} and Q is the \textit{consequent}.\\ 
    & \\ 

    \cmidrule{1-2}
    
    \multirow{2}{*}{\parbox{3cm}{\centering Deductive Reasoning \\ (\text{13.97\%})}} 
    & Deductive reasoning focuses on deriving specific \textit{conclusions} from general \textit{principles} or \textit{premises}, ensuring that \textit{conclusions} logically follow if the \textit{premises} are true. \\
    \cmidrule{1-2}
    
    \multirow{2}{*}{\parbox{3cm}{\centering Abductive Reasoning \\ (\text{16.03\%})}} 
    & Abductive reasoning, considered more \textit{creative} and \textit{open-ended}, involves forming \textit{hypotheses} to explain \textit{observations}, often generating the most \textit{plausible explanation} rather than a \textit{guaranteed conclusion}.\\

    \bottomrule
\end{tabular}}
\vspace{1ex}
\caption{Definitions and proportions of the seven subcategories in logical reasoning within \benchmark. The percentage indicates the proportion of each subcategory within the overall logical reasoning category.}
\label{tab:logical_subcategory_definition}
\end{table*}
\begin{table*}[htbp]
\centering
\small
\renewcommand\tabcolsep{2.5pt}
\resizebox{\textwidth}{!}{\begin{tabular}{@{}l p{13cm}@{}}
    \toprule
    \textbf{Numerical Reasoning} & \textbf{Description} \\
    \midrule
    
    \multirow{2}{*}{\parbox{3.5cm}{\centering Exact Number Generation \\ (\text{31.06\%})}} 
    & Exact number generation examines the model's ability to correctly generate an \textit{exact number} of \textit{objects}. \\ 
    & \\ 
    \cmidrule{1-2}
    
    \multirow{2.5}{*}{\parbox{3.5cm}{\centering Approximate Number Generation and Zero \\ (\text{31.37\%})}} 
    & Approximate number generation evaluates models on their ability to correctly depict \textit{entities} with quantities expressed in \textit{approximate terms} by means of \textit{linguistic quantifiers(e.g., "many", "a few", or  "more")}. \\
     
    \cmidrule{1-2}
    
    \multirow{3}{*}{\parbox{3.5cm}{\centering Conceptual Quantitative Reasoning \\ (\text{37.58\%})}} 
    & Conceptual quantitative reasoning evaluates models on prompts that require a \textit{conceptual understanding} of \textit{objects} and their \textit{parts}. \\
        & \\ 
    \bottomrule
\end{tabular}}
\vspace{1ex}
\caption{Definitions and proportions of the three subcategories in Numerical reasoning within \benchmark. The percentage indicates the proportion of each subcategory within the overall Numerical reasoning category.}
\label{tab:numerical_subcategory_definition}
\end{table*}

\newpage

\begin{table*}[htbp]
\centering
\small
\renewcommand\tabcolsep{2.5pt}  

\resizebox{\textwidth}{!}{\begin{tabular}{@{}l p{13cm}@{}}
    \toprule
    \textbf{Commonsense Reasoning} & \textbf{Description} \\
    \midrule
    \multirow{2}{*}{\parbox{3cm}{\centering Affordance \\ \centering(\text{14.53\%})}} 
    & Affordance commonsense reasoning involves providing a description of an object's potential \textit{use} or \textit{function}, requiring the model to generate an object based on that description. \\
    \midrule
    \multirow{2}{*}{\parbox{3cm}{\centering Attribute \\ \centering(\text{14.53\%})}} 
    & Attribute commonsense reasoning refers to the model's ability to infer or recognize the \textit{properties} and \textit{characteristics} of an object, utilizing both observable and unobservable information. \\
    \midrule
    \multirow{2}{*}{\parbox{3cm}{\centering Color \\ \centering(\text{14.82\%})}} 
    & Color commonsense reasoning pertains to the model’s ability to infer the correct \textit{color} of an object based on commonsense knowledge related to \textit{color}. \\
    \midrule
    \multirow{3}{*}{\parbox{3cm}{\centering Emotion Intention Commonsense \\ \centering(\text{11.94\%})}} 
    & Emotion intention commonsense reasoning explores the model's ability to understand \textit{emotional cues} and \textit{intentions}, particularly in the context of \textit{human-object interactions} in images. This subcategory evaluates how well the model can recognize and interpret \textit{emotional states} and \textit{intentions} from visual input. \\
    \midrule
    \multirow{2.5}{*}{\parbox{3cm}{\centering Social \& Cultural Knowledge (Object) \\
    \centering(\text{14.68\%})}} 
    & Social and cultural commonsense reasoning (Object) assesses the model’s ability to leverage knowledge related to \textit{social} and \textit{cultural contexts} when generating a specific object. \\
    & \\ 
    \midrule
    \multirow{2.5}{*}{\parbox{3cm}{\centering Social \& Cultural Knowledge (Scene) \\ \centering(\text{15.11\%})}} 
    & Social and cultural commonsense reasoning (Scene) evaluates the model’s ability to incorporate knowledge of \textit{social} and \textit{cultural contexts} when generating \textit{scenes} or \textit{environments} that accurately reflect specific \textit{social} and \textit{cultural} settings. \\
    \midrule
    \multirow{3}{*}{\parbox{3cm}{\centering Temporal Understanding \\ \centering(\text{14.39\%})}} 
    & Temporal understanding commonsense reasoning focuses on the model’s ability to infer and apply knowledge related to \textit{time-dependent changes} or \textit{events}, including the ability to predict how \textit{objects} or \textit{scenes} may evolve over time based on contextual and temporal understanding. \\
    \bottomrule
\end{tabular}}
\vspace{1ex}
\caption{Definitions and proportions of the seven subcategories in commonsense reasoning within \benchmark. The percentage indicates the proportion of each subcategory within the overall commonsense reasoning category.}
\label{tab:commosense_subcategory_definition}
\end{table*}

\newpage

\newpage

\begin{figure*}[htbp]
\centering
\includegraphics[width=\textwidth]{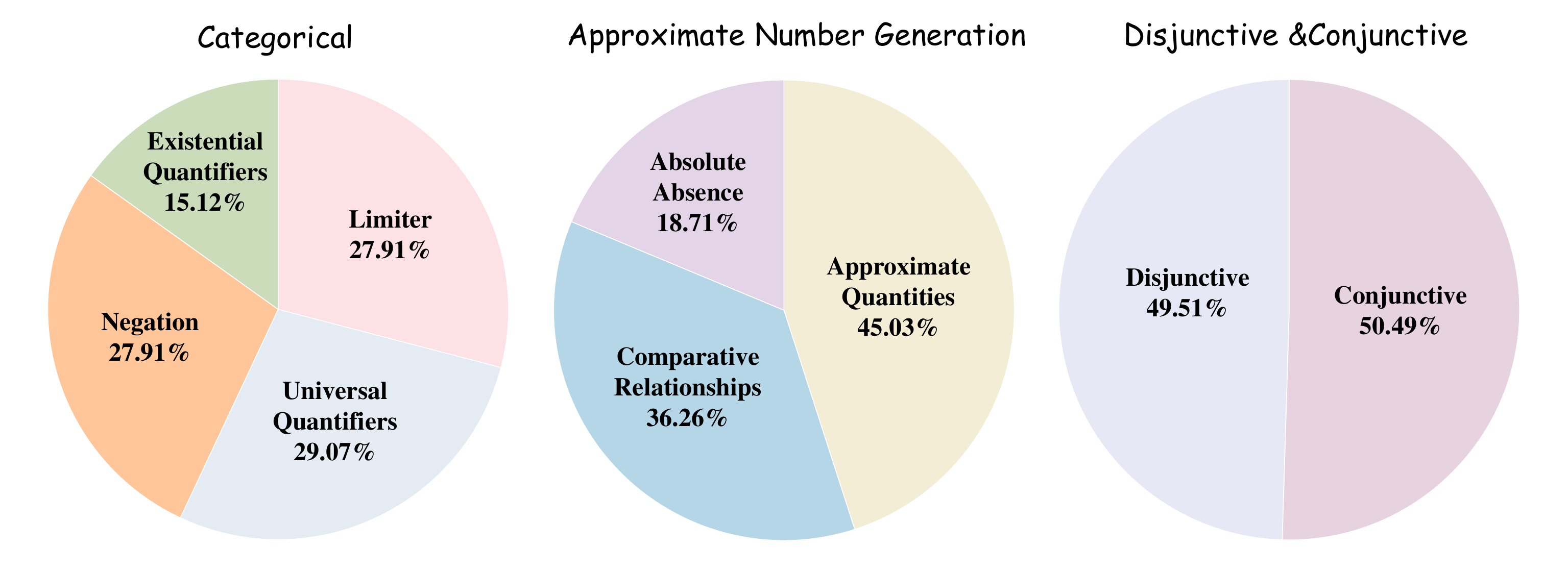}
   \caption{\textbf{Distribution of Quantifiers and Operations in Categorical, Approximate Number Generation, Disjunctive Reasoning, and Conjunctive Reasoning.}}
\label{fig:appendix_term_understanding_logical}
\end{figure*}

\begin{figure*}[htbp]
\centering
\includegraphics[width=\textwidth]{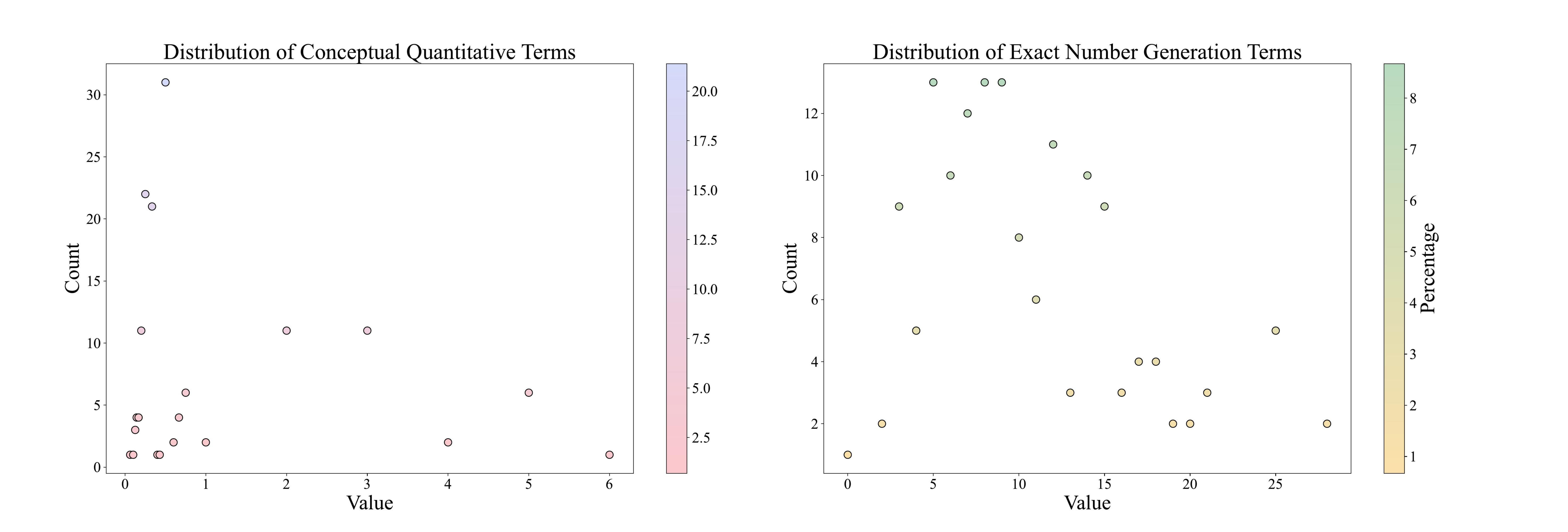}
   \caption{\textbf{Distribution of Numbers in Exact Number Generation and Conceptual Quantitative Reasoning.} Due to the current limitations of the best visual language models in numerical tasks, the numbers in Exact Number Generation are restricted to values within 30.}
\label{fig:appendix_term_understanding_numerical}
\end{figure*}

\clearpage
\section{Evaluation Details}
\label{sec:expdetails}
All experiments with open-source models are performed on A-40 GPUs, whereas experiments involving closed-source models are conducted using the API key provided by the respective service. All experiments are conducted in a zero-shot setting to assess the generalization capabilities of text-to-image (T2I) generation models on reasoning tasks, without relying on few-shot prompting or additional fine-tuning.

\subsection{Prompts details}
\label{sec:prompt_details}

\begin{figure}[htbp]
\centering
\includegraphics[width=\linewidth ]{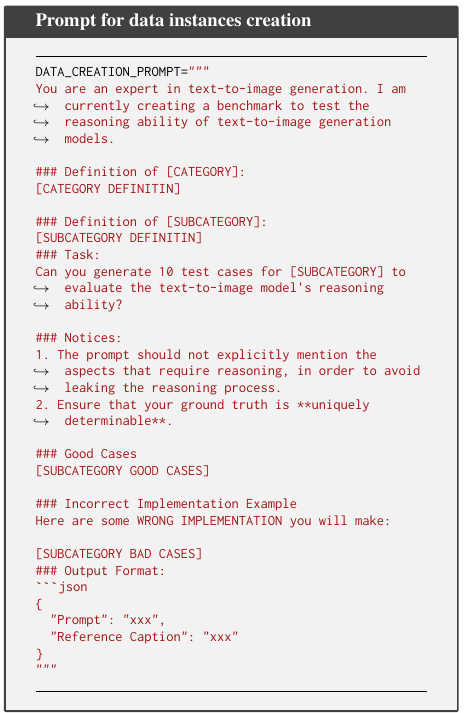}
\caption{Prompt for data instances creation}
\label{fig:prompt_for_data_instances_creation}
\end{figure}

\subsubsection{Prompt for Data Instances Creation}
\label{sec:prompt_creation}
We initially generate data instance using \modelname{gpt-4o} with the prompt in Figure~\ref{fig:prompt_for_data_instances_creation} and deleted lots of duplicated and low-quality data instances in data filtering stage.

\begin{figure}[htbp]
\centering
\includegraphics[width=\linewidth]{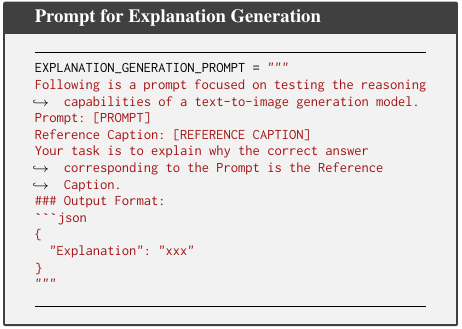}
\caption{Prompt for explanation generation} 
\label{fig:prompt_for_explanation_generation} 
\end{figure}

\subsubsection{Prompt for Explanation Generation}
\label{sec:explanation_creation}
We generate explanations using \modelname{GPT-4o} with the prompt in Figure~\ref{fig:prompt_for_explanation_generation}.

\subsubsection{Prompt for Image Evaluation}
\label{sec:promptimgevaluation}
After refining the evaluation questions and scoring criteria, we use these prompts with the state-of-the-art visual language model, GPT-4o, with a fixed temperature of 0.1, to obtain scores for each image relative to its corresponding text.

\begin{figure}[htbp]
\centering
\includegraphics[width=\linewidth]{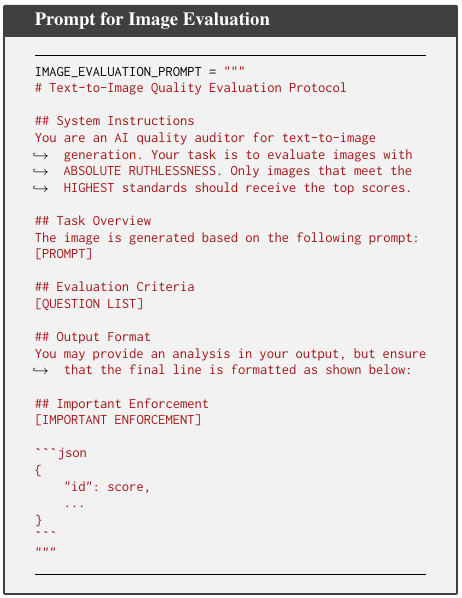}
\caption{Prompt for image evaluation} 
\label{fig:prompt_for_image_evaluation} 
\end{figure}

\begin{figure}[htbp]
\centering
\includegraphics[width=\linewidth]{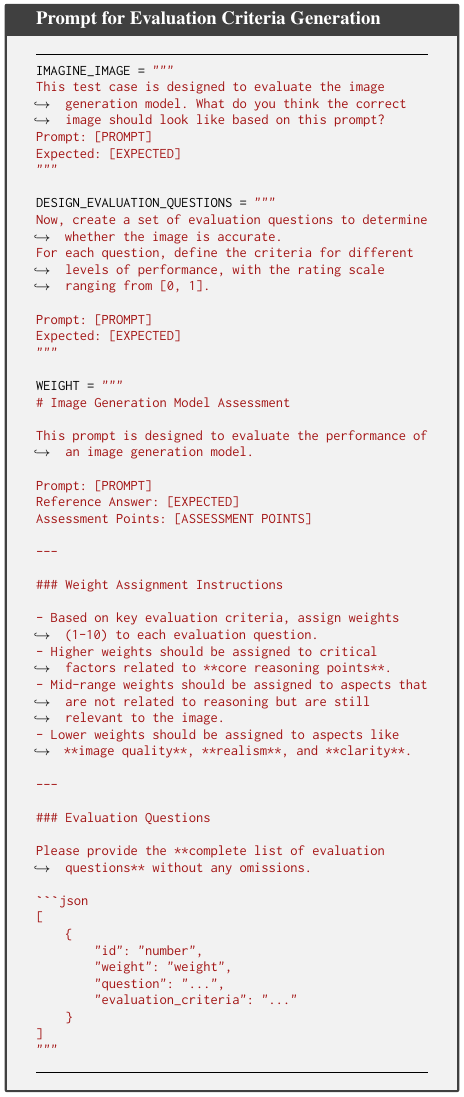}
\caption{Prompt for evaluation criteria generation} 
\label{fig:prompt_evaluation} 
\end{figure}

\subsubsection{Prompt for Evaluation Criteria Generation}
\label{sec:promptevaluationgen}

We generate evaluation questions and scoring criteria initially using the GPT-4o API with a fixed temperature of 0.1 to ensure score stability. Due to performance degradation in GPT-4o when handling long contexts, we have separated the prompts for generating questions and assigning weights. This approach ensures that GPT-4o can fully adhere to all the key points specified in each prompt. These outputs are then carefully reviewed and refined by human annotators to ensure they align with human judgment. The prompt we used is in Figure~\ref{fig:prompt_evaluation}

\subsection{Human Annotators}
\label{sec:humanannotators}
To incorporate human judgment and validate the effectiveness of our evaluation approach, we organize a group of senior college students. Each participant is tasked with comparing the image outputs generated by two similarly performing models, \modelname{Lumina-Image 2.0} and \modelname{Sana-1.5}, selecting the image they find most aligned with the prompt or indicating if both outputs are equally satisfactory or unsatisfactory. 
\subsection{Model Details}\label{appendix:model-config}
\paragraph{\textbf{Model Sources.}}
For different T2I models, we select their latest models and best-performing configurations for evaluation to fully \benchmark their reasoning ability. Table~\ref{supp-t0.5} presents the release time and model sources of MLLMs used in \benchmark.

\begin{table*}[htbp]

    \centering
    \small  
    \caption{\textbf{The Release Time and Model Source of T2I Models Evaluated in \benchmark.}}
    \resizebox{0.92\linewidth}{!}{
    \begin{tabularx}{0.96\textwidth}{@{} 
        >{\raggedright\arraybackslash}p{2.2cm} 
        >{\centering\arraybackslash}p{1.3cm} 
        >{\centering\arraybackslash}p{2cm} 
        >{\raggedright\arraybackslash}X @{}} 
    \toprule
    \textbf{Model} & \textbf{Release Time} & \textbf{Source} & \textbf{URL} \\
    \midrule
    EMU3~\cite{wang2024emu3} 
    & 2024-09 & local checkpoint 
    & \href{https://github.com/baaivision/Emu3}{\textcolor{myblue}{\url{https://github.com/baaivision/Emu3}}} \\
    \midrule
    Janus-Pro-7B~\cite{chen2025janus} & 2025-01 & local checkpoint & \textcolor{myblue}{\url{https://github.com/deepseek-ai/Janus/}} \\
    \midrule
    LlamaGen~\cite{sun2024autoregressive} & 2024-06 & local checkpoint & \textcolor{myblue}{\url{https://huggingface.co/FoundationVision/LlamaGen}} \\
    \midrule
    SD3-medium~\cite{esser2024scalingrectifiedflowtransformers} & 2024-10 & local checkpoint & \textcolor{myblue}{\url{https://huggingface.co/stabilityai/stable-diffusion-3.5-medium}} \\
    \midrule
    Lumina-Image-2.0~\cite{qin2025lumina} & 2025-03 & local checkpoint & \textcolor{myblue}{\url{https://github.com/Lumina-Image 2.0}} \\
    \midrule
    Sana-1.5~\cite{xie2025sana} & 2025-03 & local checkpoint & \textcolor{myblue}{\url{https://github.com/NVlabs/Sana}} \\
    \midrule
    Lumina-T2I~\cite{qin2025lumina} & 2024-05 & local checkpoint & \textcolor{myblue}{\url{https://huggingface.co/Alpha-VLLM/Lumina-Next-SFT-diffusers}} \\
    \midrule
    LLM4GEN\textit{$_{\text{SD1.5}}$}~\cite{liu2025llm4gen} & 2024-07 & local checkpoint & \textcolor{myblue}{\url{https://github.com/YUHANG-Ma/LLM4GEN}} \\
    \midrule
    ELLA\textit{$_{\text{SD1.5}}$}~\cite{hu2024ella} & 2024-03 & local checkpoint & \textcolor{myblue}{\url{https://github.com/TencentQQGYLab/ELLA}} \\
    \midrule
    Show-o+PARM~\cite{guo2025can} & 2025-01 & local checkpoint & \textcolor{myblue}{\url{https://huggingface.co/ZiyuG/Image-Generation-CoT}} \\
    \midrule
    Show-o+DPO~\cite{guo2025can} & 2025-01 & local checkpoint & \textcolor{myblue}{\url{https://huggingface.co/ZiyuG/Image-Generation-CoT}} \\
    \midrule
    Show-o+ORM~\cite{guo2025can} & 2025-01 & local checkpoint & \textcolor{myblue}{\url{https://huggingface.co/ZiyuG/Image-Generation-CoT}} \\
    \midrule
    gpt-image-1~\cite{hurst2024gpt} & 2025-04 & API & \textcolor{myblue}{\url{https://platform.openai.com/}} \\
    \bottomrule
    \end{tabularx}}
    \label{supp-t0.5}
\end{table*}

\section{Detailed Experimental Results}
\label{sec:detailresults}

\subsection{Main Results across 33 Subcategories}
Table ~\ref{tab:mathematicalsubcategoryresults} to~\ref{tab:conceptcompositionalsubcategoryresults} are the main results of the models across subcategories in \textit{Mathematical Reasoning, Logical Reasoning, Commonsense Reasoning, Concept Mixing Reasoning,  Causal Reasoning, Numerical Reasoning and Compositional Reasoning}.

\subsection{Comparison of Subcategory Performance: Standard T2I Model vs. Pipeline-based Framework}
\begin{figure}[htbp]
	\centering
    \vspace{-2mm}  
	\includegraphics[width=\linewidth]{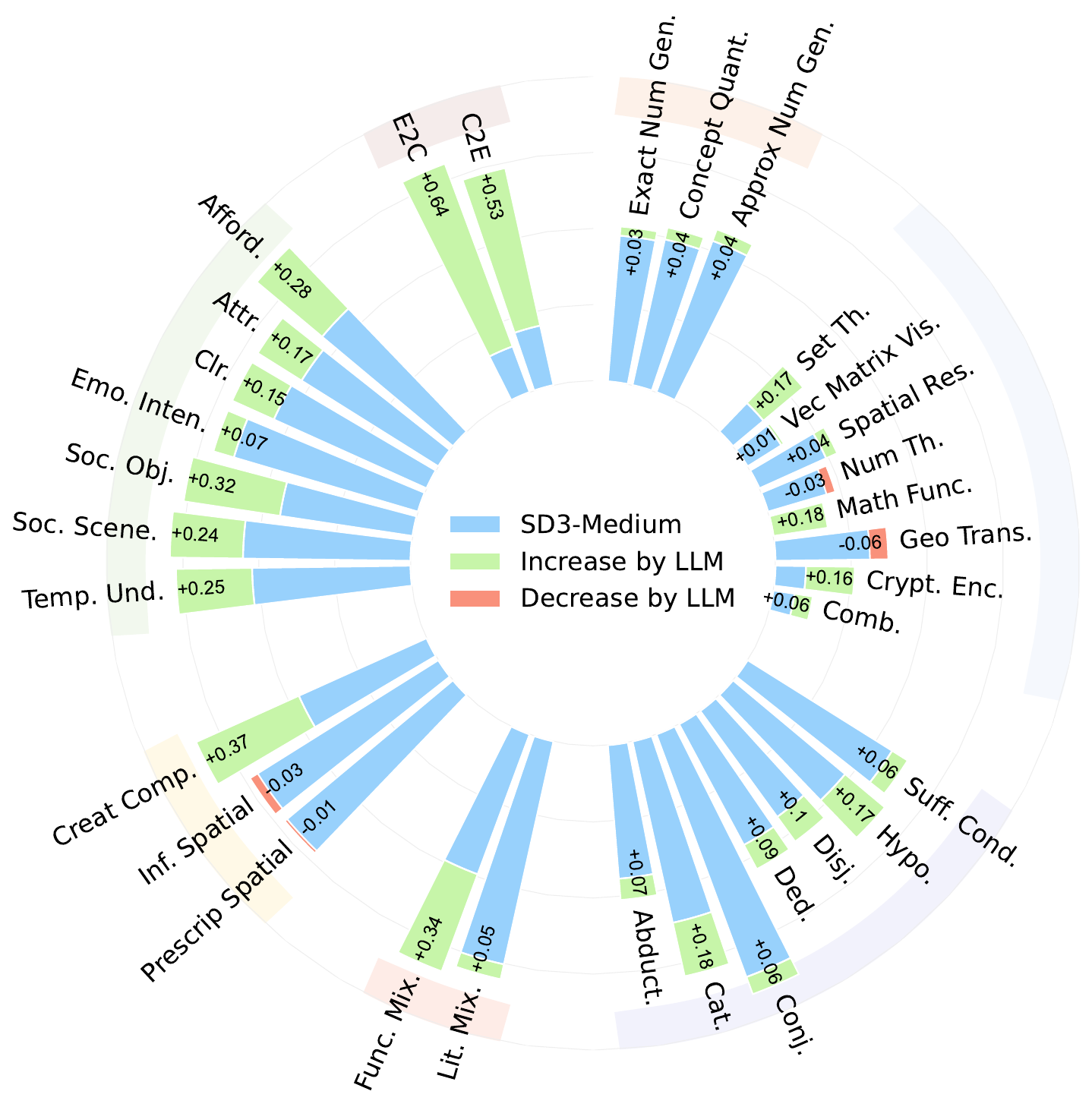}
    \vspace{-5mm}  
	\caption{\textbf{Detailed Performance Comparison: Standard T2I Model vs. Pipeline-based Framework}. We denote the results of standard T2I models in blue pillars and highlight the increase and decrease magnitude with the pipeline-based framework by green and red colors, respectively. }
\label{fig:llm-rewrite-overall-performance}
 \vspace{-3mm}
\end{figure}

Figure~\ref{fig:llm-rewrite-overall-performance} presents detailed performance comparison: standard T2I model vs. pipeline-based framework

\begin{table*}[htbp]
\centering
\small  
\begin{tabularx}{\textwidth}{
  >{\raggedright\arraybackslash}p{4.5cm}  
  *{9}{>{\centering\arraybackslash}X}  
  @{}  
}
\toprule
\textbf{Method} &  \textbf{Overall}& \textbf{Comb.} & \textbf{Crypt. Enc.} &  \textbf{Geo. Trans.} & \textbf{Math Func.} & \textbf{Num. Th.} & \textbf{Spatial Reas.} & \textbf{Vec/Mat. Vis.} & {\textbf{Set Th.}} \\
\midrule

\multicolumn{9}{c}{\textit{Diffusion Models}} \\
\midrule
SD3-medium~\cite{esser2024scaling} & 0.19 & 0.07 & 0.10 & 0.37 & 0.01 & 0.23 & 0.24 & 0.13 & 0.13 \\
Lumina-Image 2.0~\cite{qin2025lumina} & 0.13 & 0.09 & 0.09 & 0.18 & 0.03 & 0.06 & 0.28 & 0.01 & 0.16 \\
Sana-1.5~\cite{xie2025sana} & 0.13 & 0.10 & 0.06 & 0.32 & 0.02 & 0.08 & 0.16 & 0.06 & 0.16 \\
Lumina-T2I~\cite{qin2025lumina} & 0.13 & 0.04 & 0.01 & 0.17 & 0.03 & 0.07 & 0.16 & 0.05 & 0.07 \\
OminGen~\cite{xiao2024omnigen}& 0.18 & 0.19 & 0.05 & 0.27 & 0.06 & 0.33 & 0.26 & 0.08 & 0.21 \\
LLM4GEN\textit{$_{\text{SD1.5}}$}~\cite{liu2025llm4gen} & 0.07 & 0.03 & 0.01 & 0.16 & 0.01 & 0.01 & 0.01 & 0.01 & 0.09 \\
ELLA\textit{$_{\text{SD1.5}}$}~\cite{hu2024ella} & 0.07 & 0.01 & 0.03 & 0.14 & 0.01 & 0.01 & 0.11 & 0.07 & 0.07 \\
\midrule

\multicolumn{9}{c}{\textit{AutoRegressive Models}} \\
\midrule
EMU3~\cite{wang2024emu3} & 0.09 & 0.05 & 0.01 & 0.18 & 0.03 & 0.08 & 0.14 & 0.05 & 0.08 \\
Janus-Pro-7B~\cite{chen2025janus} & 0.07 & 0.02 & 0.01 & 0.16 & 0.02 & 0.06 & 0.12 & 0.01 & 0.06 \\
LlamaGen~\cite{sun2024autoregressive} & 0.07 & 0.04 & 0.01 & 0.24 & 0.01 & 0.01 & 0.01 & 0.05 & 0.10 \\
Show-o~\cite{xie2024show} & 0.12 & 0.12 & 0.02 & 0.20 & 0.01 & 0.26 & 0.19 & 0.03 & 0.14 \\

\midrule
\multicolumn{9}{c}{\textit{Reasoning-Enhanced Models}} \\
\midrule
Show-o+ORM~\cite{guo2025can} & 0.12 & 0.12 & 0.02 & 0.19 & 0.02 & 0.24 & 0.18 & 0.04 & 0.13 \\
Show-o+DPO~\cite{guo2025can} & 0.13 & 0.14 & 0.04 & 0.20 & 0.03 & 0.23 & 0.20 & 0.06 & 0.14 \\
Show-o+PARM~\cite{guo2025can} & 0.13 & 0.13 & 0.03 & 0.21 & 0.02 & 0.27 & 0.20 & 0.04 & 0.15\\
\midrule

\multicolumn{9}{c}{\textit{Close Source Models}} \\
\midrule
DALL-E3~\cite{ma2024learning} & 0.21 & 0.07 & 0.14 & 0.32 & 0.05 & 0.18 & 0.42 & 0.12 & 0.36 \\
GPT-4o~\cite{hurst2024gpt} & 0.58 & 0.43 & 0.43 & 0.73 & 0.59 & 0.49 & 0.74 & 0.46 & 0.75 \\

\bottomrule
\end{tabularx}

\vspace{1ex}
\caption{Evaluation of mathematical capabilities in generative models. Comb.: Combinatorial, Crypt. Enc.: Cryptographic Encoding, Geo. Trans.: Geometrical Transformations, Math Func.: Mathematical Function, Num. Th.: Number Theory, Spatial Reas.: Spatial Reasoning, Vec/Mat. Vis.: Vector \& Matrix Visualization, Set Th.: Set Theory.}

\label{tab:mathematicalsubcategoryresults}
\end{table*}

\begin{table*}[htbp]
\centering
\small  
\begin{tabularx}{\textwidth}{
  >{\raggedright\arraybackslash}p{4.5cm}  
  *{10}{>{\centering\arraybackslash}X}  
  @{}  
}
\toprule
\textbf{Method} & \textbf{Overall} & \textbf{Abduc.} & \textbf{Cat.} & \textbf{Conj.} & \textbf{Ded.} &\textbf{Disj.} & \textbf{Hypo.} & \textbf{Suff.} \\
\midrule
\multicolumn{9}{c}{\textit{Diffusion Models}} \\
\midrule
SD3-medium~\cite{esser2024scaling} & 0.55 & 0.44 & 0.61 & 0.85 & 0.45 & 0.43 & 0.48 & 0.56 \\
Lumina-Image 2.0~\cite{qin2025lumina} & 0.56 & 0.44 & 0.57 & 0.87 & 0.38 & 0.51 & 0.52 & 0.54 \\
Sana-1.5~\cite{xie2025sana} & 0.49 & 0.46 & 0.56 & 0.89 & 0.50 & 0.48 & 0.59 & 0.56 \\
Lumina-T2I~\cite{qin2025lumina} & 0.38 & 0.33 & 0.50 & 0.69 & 0.54 & 0.43 & 0.55 & 0.53 \\
OminGen~\cite{xiao2024omnigen} & 0.51 & 0.42 & 0.64 & 0.69 & 0.39 & 0.52 & 0.41 & 0.47 \\
LLM4GEN\textit{$_{\text{SD1.5}}$}~\cite{liu2025llm4gen} & 0.55 & 0.33 & 0.48 & 0.70 & 0.40 & 0.53 & 0.55 & 0.49 \\
ELLA\textit{$_{\text{SD1.5}}$}~\cite{hu2024ella} & 0.40 & 0.29 & 0.41 & 0.64 & 0.26 & 0.59 & 0.40 & 0.39 \\
\midrule
\multicolumn{9}{c}{\textit{AutoRegressive Models}} \\
\midrule
EMU3~\cite{wang2024emu3} & 0.55 & 0.38 & 0.53 & 0.71 & 0.44 & 0.64 & 0.52 & 0.58 \\
Janus-Pro-7B~\cite{chen2025janus} & 0.46 & 0.25 & 0.64 & 0.85 & 0.13 & 0.57 & 0.52 & 0.22 \\
LlamaGen~\cite{sun2024autoregressive} & 0.38 & 0.15 & 0.48 & 0.55 & 0.17 & 0.59 & 0.29 & 0.35 \\
Show-o~\cite{xie2024show} & 0.42 & 0.40 & 0.62 & 0.71 & 0.35 & 0.42 & 0.32 & 0.38 \\
\midrule

\multicolumn{9}{c}{\textit{Reasoning-Enhanced Models}} \\
\midrule
Show-o+ORM~\cite{guo2025can} & 0.37 & 0.33 & 0.48 & 0.47 & 0.35 & 0.41 & 0.37 & 0.17 \\
Show-o+DPO~\cite{guo2025can} & 0.41 & 0.29 & 0.44 & 0.44 & 0.36 & 0.43 & 0.34 & 0.18 \\
Show-o+PARM~\cite{guo2025can} & 0.45 & 0.38 & 0.53 & 0.76 & 0.33 & 0.45 & 0.31 & 0.36 \\
\midrule
\multicolumn{9}{c}{\textit{Close Source Models}} \\
\midrule
DALLE~\cite{ma2024learning} & 0.69 & 0.56 & 0.67 & 0.87 & 0.70  & 0.46 & 0.79 & 0.78 \\
gpt-image-1~\cite{ma2024learning} & 0.81 & 0.79  & 0.88  & 0.95  & 0.79   & 0.76 &  0.79 & 0.73 \\
\bottomrule
\end{tabularx}
\vspace{1ex}
\caption{Evaluation of text-to-image generation on Logical Reasoning in \benchmark. Abduc.: Abductive, Cat.: Categorical, Conj.: Conjunctive, Ded.: Deductive, Disj.: Disjunctive, Hypo.: Hypothetical, Suff.: Sufficient Conditional}
\label{tab:logicalsubcategoryresults}
\end{table*}

\begin{table*}[htbp]
\centering
\small  
\begin{tabularx}{\textwidth}{
  >{\raggedright\arraybackslash}p{4.5cm}  
  *{8}{>{\centering\arraybackslash}X}  
  @{}  
}
\toprule
\textbf{Method} &\textbf{Overall} &\textbf{Afford.} &\textbf{Attribute} &\textbf{Color} &\textbf{Emotion} &\textbf{Object} &\textbf{Scene} & \textbf{Temp.} \\
\midrule
\multicolumn{9}{c}{\textit{Diffusion Models}} \\
\midrule
SD3-medium~\cite{esser2024scaling} & 0.54 & 0.56 & 0.53 & 0.55 & 0.63 & 0.44 & 0.55 & 0.52 \\
Lumina-Image 2.0~\cite{qin2025lumina} & 0.49 & 0.46 & 0.53 & 0.51 & 0.65 & 0.34 & 0.53 & 0.46 \\
Sana-1.5~\cite{xie2025sana} & 0.49 & 0.42 & 0.60 & 0.51 & 0.64 & 0.33 & 0.53 & 0.51 \\
Lumina-T2I~\cite{qin2025lumina} & 0.38 & 0.36 & 0.47 & 0.40 & 0.57 & 0.33 & 0.46 & 0.39 \\
OminGen~\cite{xiao2024omnigen} & 0.43 & 0.41 & 0.51 & 0.39 & 0.54 & 0.30 & 0.47 & 0.41 \\ 
LLM4GEN\textit{$_{\text{SD1.5}}$}~\cite{liu2025llm4gen} & 0.55 & 0.37 & 0.47 & 0.44 & 0.66 & 0.36 & 0.56 & 0.51 \\
ELLA\textit{$_{\text{SD1.5}}$}~\cite{hu2024ella} & 0.40 & 0.33 & 0.40 & 0.34 & 0.37 & 0.28 & 0.36 & 0.32 \\
\midrule
\multicolumn{9}{c}{\textit{AutoRegressive Models}} \\
\midrule
EMU3~\cite{wang2024emu3} & 0.46 & 0.40 & 0.50 & 0.43 & 0.58 & 0.39 & 0.52 & 0.42 \\
Janus-Pro-7B~\cite{chen2025janus} & 0.45 & 0.38 & 0.57 & 0.45 & 0.58 & 0.32 & 0.49 & 0.40 \\
LlamaGen~\cite{sun2024autoregressive} & 0.38 & 0.38 & 0.42 & 0.39 & 0.40 & 0.29 & 0.38 & 0.38 \\
Show-o~\cite{xie2024show} & 0.42 & 0.44 & 0.48 & 0.41 & 0.44 & 0.32 & 0.44 & 0.36 \\
\midrule
\multicolumn{9}{c}{\textit{Reasoning-Enhanced Models}} \\
\midrule
Show-o+ORM~\cite{guo2025can} & 0.42 & 0.42 & 0.49 & 0.40 & 0.47 & 0.35 & 0.47 & 0.38\\
Show-o+DPO~\cite{guo2025can} & 0.43 & 0.43 & 0.52 & 0.44 & 0.45 & 0.36 & 0.47 & 0.36\\
Showo-o+PARM~\cite{guo2025can} & 0.45 & 0.45 & 0.48 & 0.46 & 0.55 & 0.40 & 0.49 & 0.47\\
\midrule
\multicolumn{9}{c}{\textit{Close Source Models}} \\
\midrule
DALLE3~\cite{ma2024learning} & 0.78 & 0.70 & 0.80 & 0.86 & 0.81 & 0.81 & 0.77 & 0.72 \\
gpt-iamge-1~\cite{ma2024learning} & 0.83 & 0.89 & 0.79 & 0.80 & 0.89 &  0.85 & 0.87 & 0.75 \\
\bottomrule
\end{tabularx}

\vspace{1ex}
\caption{Evaluation Results of text-to-image generation on Commonsense Reasoning in \benchmark. Afford.: Affordance. Temp.: Temporal Understanding. Emotion: Emotion Intention Commonsense Reasoning. Object: Social Cultural Knowledge (Object). Scene: Social Cultural Knowledge (Scene). }
\label{tab:commonsensesubcategoryresults}
\end{table*}

\begin{table*}[htbp]
\resizebox{\linewidth}{!}{
\begin{tabular}{l|c|ccc|c|cc}
\toprule
\multirow{2}{*}{\makecell*[c]{\textbf{Method}}} 
& \multirow{2}{*}{\makecell*[c]{\textbf{Overall}}} 
& \multicolumn{3}{c|}{\textbf{Numerical}} 
& \multirow{2}{*}{\textbf{Overall}} 
& \multicolumn{2}{c}{\textbf{Causal Reasoning}} \\
\cmidrule(lr){3-5} \cmidrule(lr){7-8}  
&  
& \textbf{Approx.} 
& \textbf{Conceptual.} 
& \textbf{Exact.} 
&  
& \textbf{C2E} 
& \textbf{E2C} \\
\midrule
\multicolumn{8}{c}{\textit{Diffusion Models}} \\
\midrule
SD3-medium~\cite{esser2024scaling} & 0.50 & 0.53 & 0.49 & 0.48 & 0.18 & 0.20 & 0.16 \\
Lumina-Image 2.0~\cite{qin2025lumina} & 0.43 & 0.54 & 0.40 & 0.35 & 0.40 & 0.37 & 0.44 \\
Sana-1.5~\cite{xie2025sana} & 0.47 & 0.58 & 0.37 & 0.47 & 0.21 & 0.23 & 0.19 \\
Lumina-T2I~\cite{qin2025lumina} & 0.45 & 0.53 & 0.45 & 0.38 & 0.18 & 0.18 & 0.18 \\
OminGen~\cite{xiao2024omnigen} & 0.47 & 0.59 & 0.40 & 0.42 & 0.34 & 0.26 & 0.41 \\
LLM4GEN\textit{$_{\text{SD1.5}}$}~\cite{liu2025llm4gen} & 0.39 & 0.44 & 0.36 & 0.36 & 0.45 & 0.46 & 0.44 \\
ELLA\textit{$_{\text{SD1.5}}$}~\cite{hu2024ella} & 0.32 & 0.41 & 0.25 & 0.30 & 0.29 & 0.22 & 0.38 \\
\midrule

\multicolumn{8}{c}{\textit{AutoRegressive Models}} \\
\midrule
EMU3~\cite{wang2024emu3} & 0.61 & 0.73 & 0.54 & 0.56 & 0.41 & 0.36 & 0.47 \\
Janus-Pro-7B~\cite{chen2025janus} & 0.46 & 0.53 & 0.38 & 0.48 & 0.36 & 0.34 & 0.39 \\
LlamaGen~\cite{sun2024autoregressive} & 0.35 & 0.43 & 0.31 & 0.30 & 0.12 & 0.12 & 0.12 \\
Show-o~\cite{xie2024show} & 0.57 & 0.68 & 0.50 & 0.53 & 0.30 & 0.23 & 0.38 \\
\midrule

\multicolumn{8}{c}{\textit{Reasoning-Enhanced Models}} \\
\midrule
Show-o+ORM~\cite{guo2025can} & 0.49 & 0.52 & 0.46 & 0.49 & 0.26 & 0.30 & 0.23 \\
Show-o+DPO~\cite{guo2025can} & 0.51 & 0.58 & 0.46 & 0.50 & 0.31 & 0.35 & 0.28 \\
Show-o+PARM~\cite{guo2025can} & 0.56 & 0.65 & 0.49 & 0.53 & 0.32 & 0.36 & 0.27 \\
\midrule
\multicolumn{8}{c}{\textit{Close Source Models}} \\
\midrule
DALLE~\cite{ma2024learning} & 0.69 & 0.71 & 0.64 & 0.72 & 0.64 & 0.69 & 0.59 \\
GPT-4o~\cite{hurst2024gpt} & 0.88 & 0.90 & 0.81 & 0.92 & 0.71 & 0.85 & 0.56 \\
\bottomrule

\end{tabular}
}
\vspace{1ex} 
\caption{Evaluation of text-to-image generation on Numerical Reasoning and Causal Reasoning in \benchmark. Approx.: Approximate Number Generation. Conceptual: Conceptual Quantitative Reasoning. Exact: Exact Number Generation. C2E: Cause to Effect Reasoning. E2C: Effect to Cause Reasoning. }
\label{tab:numericalcausalsubcategoryresults}
\end{table*}

\begin{table*}[htbp]
\resizebox{\linewidth}{!}{
\begin{tabular}{l|c|cc|c|ccc}
\toprule
\multirow{2}{*}{\textbf{Method}} 
& \multicolumn{1}{c|}{\multirow{2}{*}{\textbf{Overall}}} 
& \multicolumn{2}{c|}{\textbf{Concept Mixing}} 
& \multirow{2}{*}{\textbf{Overall}} 
& \multicolumn{3}{c}{\textbf{Compositional}} \\
\cmidrule(lr){3-4} \cmidrule(lr){6-8}  
&  
& \textbf{Functional} 
& \textbf{Literal} 
& 
& \textbf{Creative} 
& \textbf{Inferential} 
& \textbf{Prescriptive} \\
\midrule

\multicolumn{7}{c}{\textit{Diffusion Models}} \\
\midrule
SD3-medium~\cite{esser2024scaling} & 0.63 & 0.49 & 0.75 & 0.64 & 0.46 & 0.73 & 0.72 \\
Lumina-Image 2.0~\cite{qin2025lumina} & 0.54 & 0.52 & 0.56 & 0.65 & 0.50 & 0.72 & 0.73 \\
Sana-1.5~\cite{xie2025sana} & 0.66 & 0.55 & 0.75 & 0.67 & 0.59 & 0.79 & 0.63 \\
Lumina-T2I~\cite{qin2025lumina} & 0.55 & 0.47 & 0.62 & 0.49 & 0.42 & 0.56 & 0.49 \\
Omnigen~\cite{xiao2024omnigen} & 0.43 & 0.27 & 0.58 & 0.60 & 0.46 & 0.80 & 0.54 \\
LLM4GEN\textit{$_{\text{SD1.5}}$}~\cite{liu2025llm4gen} & 0.60 & 0.48 & 0.70 & 0.48 & 0.44 & 0.61 & 0.39 \\
ELLA\textit{$_{\text{SD1.5}}$}~\cite{hu2024ella} & 0.40 & 0.33 & 0.46 & 0.44 & 0.34 & 0.55 & 0.43 \\
\midrule

\multicolumn{7}{c}{\textit{AutoRegressive Models}} \\
\midrule
EMU3~\cite{wang2024emu3} & 0.62 & 0.51 & 0.70 & 0.59 & 0.50 & 0.68 & 0.59 \\
Janus-Pro-7B~\cite{chen2025janus} & 0.64 & 0.55 & 0.71 & 0.60 & 0.56 & 0.73 & 0.52 \\
LlamaGen~\cite{sun2024autoregressive} & 0.49 & 0.45 & 0.53 & 0.39 & 0.42 & 0.50 & 0.27 \\
Show-o~\cite{xie2024show} & 0.56 & 0.42 & 0.68 & 0.55 & 0.41 & 0.65 & 0.60 \\
\midrule

\multicolumn{7}{c}{\textit{Reasoning-Enhanced Models}} \\
\midrule
ORM~\cite{guo2025can} & 0.44 & 0.30 & 0.56 & 0.45 & 0.35 & 0.54 & 0.45 \\
DPO~\cite{guo2025can} & 0.48 & 0.35 & 0.61 & 0.47 & 0.38 & 0.56 & 0.47 \\
PARM~\cite{guo2025can} & 0.51 & 0.37 & 0.63 & 0.49 & 0.39 & 0.58 & 0.51 \\
\midrule

\multicolumn{7}{c}{\textit{Close Source Models}} \\
\midrule
DALLE3~\cite{ma2024learning} & 0.86 & 0.82 & 0.90 & 0.76 & 0.73 & 0.82 & 0.72 \\
GPT-4o~\cite{hurst2024gpt} & 0.89 & 0.88 & 0.90 & 0.87 & 0.81 & 0.84 & 0.95 \\
\bottomrule
\end{tabular}
}
\vspace{1ex}
\caption{Evaluation of text-to-image generation on Concept Mixing and Compositional Reasoning in \benchmark. Functional: Functional Mixing Reasoning. Literal: Literal Mixing Reasoning. Creative: Creative Compositional Reasoning. Inferential: Inferential Spatial Reasoning. Prescriptive: Prescriptive Spatial Reasoning}
\label{tab:conceptcompositionalsubcategoryresults}
\end{table*}

\subsection{Results of Our Evaluation Methods and Additional Metrics in the Benchmark}
\label{sec:appendix_metric}

In this section, we present the results of the evaluation methods employed, along with other metrics. The detailed evaluation results are provided in Table~\ref{tab:detailmetric}.

\begin{table*}[htbp]
\centering
\footnotesize
\caption{Comparison of our evaluation methods and other image-text alignment metrics across different models and categories.}
\label{tab:bias}
\resizebox{\linewidth}{!}{
\begin{tabular}{llccc}
\toprule
\textbf{Category} & \textbf{Models} & \textbf{Pairwise Accuracy $\uparrow$} & \textbf{Kendall's $\tau$ $\uparrow$} & \textbf{Spearman’s Rank Correlation $\uparrow$}  \\
\midrule
\multirow{5}{*}{Commonsense} 
& CLIPScore~\cite{hessel2021clipscore} & 0.61 & 0.22 & 0.42  \\
& DSGScore~\cite{cho2023davidsonian} & 0.54 & 0.10 & 0.30  \\
& VIEScore~\cite{ku2024viescoreexplainablemetricsconditional} & 0.70 & 0.45 & 0.34  \\
& VQAscore~\cite{lin2024evaluating} & 0.60 & 0.22 & 0.39  \\
& Ours & 0.64 & 0.60 & 0.62  \\
\midrule
\multirow{5}{*}{Compositional} 
& CLIPScore~\cite{hessel2021clipscore} & 0.71 & 0.42 & 0.39 \\
& DSGScore~\cite{cho2023davidsonian} & 0.50 & 0.38 & 0.26  \\
& VIEScore~\cite{ku2024viescoreexplainablemetricsconditional} & 0.58 & 0.40 & 0.32  \\
& VQAscore~\cite{lin2024evaluating} & 0.64 & 0.48 & 0.45  \\
& Ours & 0.73 & 0.76 & 0.61  \\
\midrule
\multirow{5}{*}{Logical} 
& CLIPScore~\cite{hessel2021clipscore} & 0.61 & 0.22 & 0.30  \\
& DSGScore~\cite{cho2023davidsonian} & 0.63 & 0.15 & 0.25  \\
& VIEScore~\cite{ku2024viescoreexplainablemetricsconditional} & 0.78 & 0.63 & 0.40  \\
& VQAscore~\cite{lin2024evaluating} & 0.76 & 0.72 & 0.68  \\
& Ours & 0.76 & 0.72 & 0.63  \\
\midrule
\multirow{5}{*}{Causal} 
& CLIPScore~\cite{hessel2021clipscore} & 0.54 & 0.18 & 0.21  \\
& DSGScore~\cite{cho2023davidsonian} & 0.51 & 0.22 & 0.28  \\
& VIEScore~\cite{ku2024viescoreexplainablemetricsconditional} & 0.69 & 0.64 & 0.68  \\
& VQAscore~\cite{lin2024evaluating} & 0.62 & 0.33 & 0.70  \\
& Ours & 0.69 & 0.64 & 0.64  \\
\midrule
\multirow{5}{*}{Concept Mixing} 
& CLIPScore~\cite{hessel2021clipscore} & 0.62 & 0.24 & 0.25  \\
& DSGScore~\cite{cho2023davidsonian} & 0.42 & 0.25 & 0.18  \\
& VIEScore~\cite{ku2024viescoreexplainablemetricsconditional} & 0.52 & 0.16 & 0.28  \\
& VQAscore~\cite{lin2024evaluating} & 0.67 & 0.52 & 0.48  \\
& Ours & 0.83 & 0.91 & 0.87  \\
\midrule
\multirow{5}{*}{Numerical} 
& CLIPScore~\cite{hessel2021clipscore} & 0.61 & 0.22 & 0.16  \\
& DSGScore~\cite{cho2023davidsonian} & 0.47 & 0.21 & 0.29  \\
& VIEScore~\cite{ku2024viescoreexplainablemetricsconditional} & 0.87 & 0.74 & 0.68  \\
& VQAscore~\cite{lin2024evaluating} & 0.78 & 0.64 & 0.57  \\
& Ours & 0.65 & 0.67 & 0.62  \\
\midrule
\multirow{5}{*}{Mathematical} 
& CLIPScore~\cite{hessel2021clipscore} & 0.72 & 0.44 & 0.54  \\
& DSGScore~\cite{cho2023davidsonian} & 0.60 & 0.45 & 0.43  \\
& VIEScore~\cite{ku2024viescoreexplainablemetricsconditional} & 0.72 & 0.44 & 0.46  \\
& VQAscore~\cite{lin2024evaluating} & 0.63 & 0.33 & 0.67  \\
& Ours & 0.69 & 0.93 & 0.87  \\
\midrule
\multirow{5}{*}{Average} 
& CLIPScore~\cite{hessel2021clipscore} & 0.631 & 0.263 & 0.310  \\
& DSGScore~\cite{cho2023davidsonian} & 0.520 & 0.220 & 0.254  \\
& VIEScore~\cite{ku2024viescoreexplainablemetricsconditional} & 0.694 & 0.494 & 0.451  \\
& VQAscore~\cite{lin2024evaluating} & 0.629 & 0.463 & 0.563  \\
& Ours & 0.713 & 0.747 & 0.694  \\

\bottomrule
\end{tabular}
}
\label{tab:detailmetric}
\end{table*}

\end{document}